\documentclass{article} 
\usepackage{iclr2024_conference,times}
\iclrfinalcopy
 
\usepackage{amsmath,amsfonts,bm}

\def\eqref#1{equation~\ref{#1}}

\def\1{\bm{1}}

\DeclareMathAlphabet{\mathsfit}{\encodingdefault}{\sfdefault}{m}{sl}
\SetMathAlphabet{\mathsfit}{bold}{\encodingdefault}{\sfdefault}{bx}{n}

\DeclareMathOperator*{\argmax}{arg\,max}

\definecolor{citecolor}{HTML}{0071bc}
\definecolor{camera}{rgb}{1.0, 0.0, 0.0}
\definecolor{revision}{rgb}{0, 0.7, 0.3}
\usepackage[breaklinks=true,colorlinks,citecolor=citecolor]{hyperref}       
\usepackage{url}

\usepackage{booktabs}       

\usepackage{algorithm}
\usepackage{algpseudocode}

\usepackage{graphicx}[pdftex]
\usepackage{subfigure}
\usepackage{amssymb}
\usepackage{adjustbox}
\usepackage{tabularx}
\usepackage{xspace}
\usepackage{wrapfig}
\usepackage[frozencache,cachedir=.]{minted}
\usepackage{enumitem}
\usepackage{multirow}

\usepackage{pifont}

\def \name{\textsc{SuRe}\xspace}

\title{
\name{}: Summarizing Retrievals using Answer\\ Candidates for Open-domain QA of LLMs}
\author{
Jaehyung Kim$^{1,}$\thanks{~~This work is done when Jaehyung Kim was at KAIST.}~~~~Jaehyun Nam$^{2}$~~ Sangwoo Mo$^{3}$~~~Jongjin Park$^{2}$ \\ \textbf{~Sang-Woo Lee$^{2,5}$~~~Minjoon Seo$^{2}$~~~Jung-Woo Ha$^{4,5}$~~Jinwoo Shin$^{2}$} \\ 
$^{1}$Carnegie Mellon University $^{2}$KAIST AI
$^{3}$University of Michigan $^{4}$Naver AI Lab $^{5}$Naver Cloud \\
\texttt{jaehyun4@andrew.cmu.edu}
}

\begin{document}

\maketitle

\begin{abstract}
Large language models (LLMs) have made significant advancements in various natural language processing tasks, including question answering (QA) tasks. 
While incorporating new information with the retrieval of relevant passages is a promising way to improve QA with LLMs, the existing methods often require additional fine-tuning which becomes infeasible with recent LLMs. 
Augmenting retrieved passages via prompting has the potential to address this limitation, but this direction has been limitedly explored.
To this end, we design a simple yet effective framework to enhance open-domain QA (ODQA) with LLMs, based on the summarized retrieval (\name{}).
\name{} helps LLMs predict more accurate answers for a given question, which are well-supported by the summarized retrieval that could be viewed as an explicit rationale extracted from the retrieved passages. 
Specifically, \name{} first constructs summaries of the retrieved passages for each of the multiple answer candidates. 
Then, \name{} confirms the most plausible answer from the candidate set by evaluating the validity and ranking of the generated summaries.
Experimental results on diverse ODQA benchmarks demonstrate the superiority of \name{}, with improvements of up to 4.6\% in exact match (EM) and 4.0\% in F1 score over standard prompting approaches. 
\name{} also can be integrated with a broad range of retrieval methods and LLMs. 
Finally, the generated summaries from \name{} show additional advantages to measure the importance of retrieved passages and serve as more preferred rationales by models and humans.\footnote{The code is available at \url{https://github.com/bbuing9/ICLR24_SuRe}}
\end{abstract}

\section{Introduction}

Large language models (LLMs)~\citep{brown2020language, touvron2023llama} have significantly accelerated progress in natural language processing (NLP) and have become a core technology in various real-world applications used by millions of users, such as coding assistants~\citep{chen2021evaluating}, search engines~\citep{xuan2023evaluation}, and chatbots~\citep{kim2021changes, openai2022chatgpt}. 
However, LLMs often suffer from limitations, such as non-factual but seemingly plausible generation, referred to as hallucinations \citep{welleck2020neural}, and difficulty in integrating up-to-date knowledge, as their learned knowledge is limited by the training corpus encoded in their parameters \citep{guu2020retrieval}. 
This problem is particularly critical for question answering (QA) \citep{kwiatkowski2019natural}, one of the most frequently encountered applications for LLMs.

Incorporating new information through the retrieval of relevant knowledge for a given query (\textit{e.g.}, a question from users) is widely explored to improve the accuracy of QA systems, called open-domain QA (ODQA) \citep{karpukhin2020dense}, and shows promise in addressing the aforementioned limitations of LLMs \citep{mialon2023augmented}.
Constructing these \textit{retrieval-augmented} LLMs typically involves
additional fine-tuning \citep{ borgeaud2022improving, izacard2022few}, but it becomes infeasible due to the increase in scale and the recent nature of black-box API \citep{openai2023gpt4}. 
Consequently, retrieval augmentation via \textit{prompting}, \textit{i.e.}, giving specific instruction as the input to obtain the desired outputs by LLM, becomes an attractive direction from its simplicity and efficiency \citep{shi2023replug}. 
However, na\"ive prompting could be limited in fully exploiting the retrieved contexts, since LLMs are simply instructed to use the retrieved information, instead of being explicitly trained to use it; for example, \citet{liu2023lost} recently observed that LLMs struggle to handle long input contexts when they are na\"ively appended. 
Despite its importance, how to improve retrieval-augmented LLMs via prompting has been under-explored.
Therefore, to improve ODQA via LLMs, we aim to develop a simple yet effective framework based on prompting, that could be easily applicable to various LLMs and retrieval methods. 

\textbf{Contribution.} 
We propose a framework based on \textbf{Su}mmarized \textbf{Re}trieval (\textbf{\name{}}), to improve ODQA performance of retrieval-augmented LLMs. 
At a high level, \name{} helps LLMs predict more grounded answers, which are well-supported by the summarization of retrieved passages that could be viewed as an explicit rationale extracted from the retrieved passages.   
To be specific, \name{} first constructs the multiple summarizations of retrieved passages conditioned on each of a few possible answer candidates. 
It enables LLMs to focus on the specific contexts relevant to the given candidate, and hence provides more discriminative viewpoints for the given question.
Then, using the generated summarizations, \name{} confirms the most plausible answer among candidates by measuring the corresponding summaries' validity to support the given candidate and ranking of relative informativeness to answer the question. 
Remarkably, all the procedures of \name{} are conducted via \textit{zero-shot prompting}. 
Consequently, \name{} is widely applicable when LLMs are only accessible with black-box API, even without query-relevant few-shot examples.

\begin{wrapfigure}[20]{r}{0.45\textwidth}
	\vspace{-0.5cm}
	{
	\includegraphics[width=60mm]{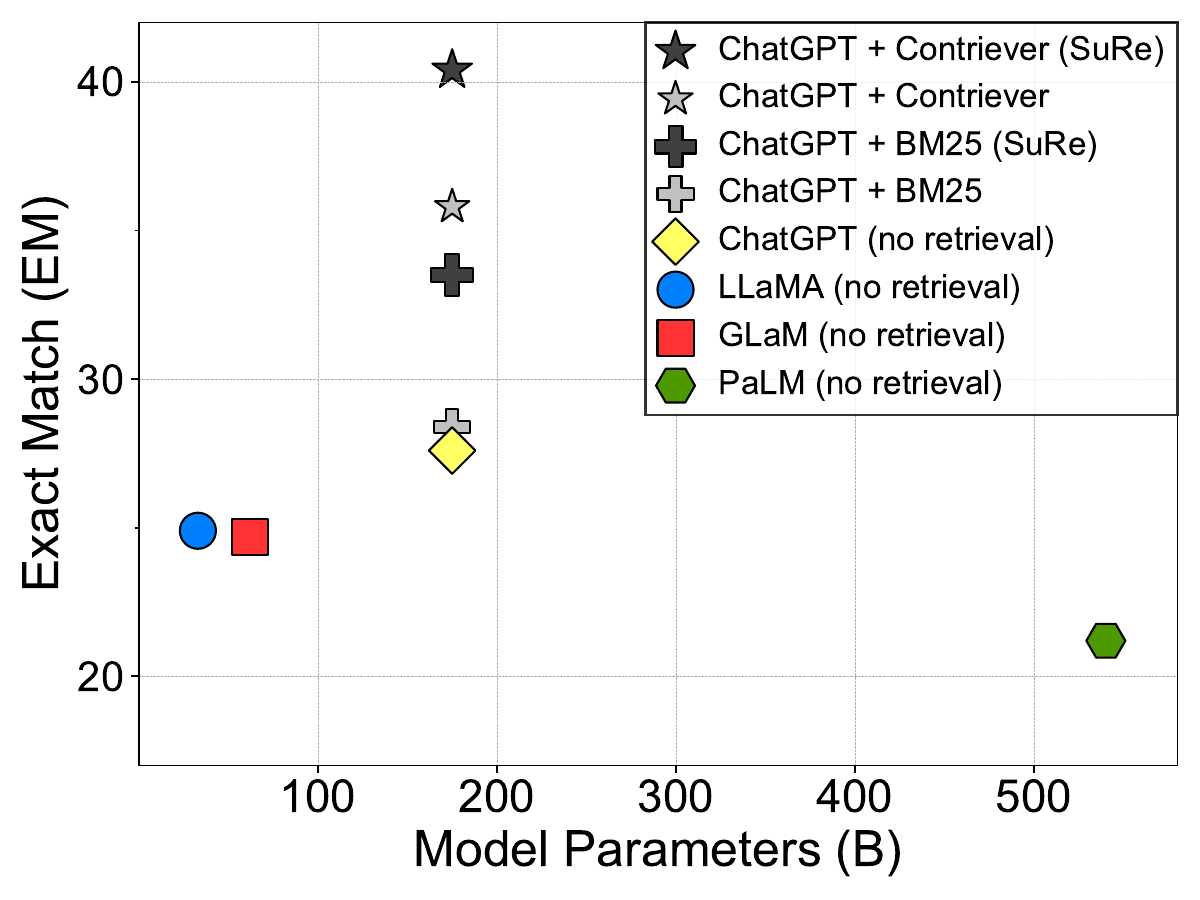}
	\vspace{-0.2cm}
        \caption{Zero-shot QA accuracy with various LLMs on Natural Question \citep{kwiatkowski2019natural}. The performances of LLaMA-33B, GLaM-62B, and PaLM-540B are from the corresponding papers, respectively \citep{chowdhery2022palm, du2022glam, touvron2023llama1}.}
 \label{common_lm}
	}
\end{wrapfigure}
Through the experiments on four different QA datasets, we demonstrate the effectiveness of \name{} for improving the zero-shot ODQA performance of retrieval-augmented LLMs.
For example, we observe that the augmentation of 10 relevant passages effectively improves QA accuracy (up to 8.2\% with Contriver \citep{izacard2021unsupervised}) of ChatGPT \citep{openai2022chatgpt}, and the gain is significantly enlarged with \name{} (up to 12.8\%), as shown in Figure \ref{common_lm}. 
Overall, \name{} with ChatGPT and BM25 \citep{robertson2009probabilistic} exhibited 4.6\%/4.0\% exact match (EM)/F1 score improvements compared to the standard prompting in average on four ODQA datasets.
In addition, \name{} is well generalized to different configurations of various retrieval methods and LLMs.
More interestingly, we observe that the generated summarization by \name{} could be further utilized to evaluate the importance of the retrieved passages, and also verify that it has a higher model/human preference as a rationale for the given prediction, compared to the generic summarization of retrieved passages.
Overall, these results highlight the effectiveness of \name{}, to improve ODQA systems based on LLMs, not only in terms of accuracy but also of additional advantages that can improve the user experience.
We, therefore, hope that the proposed framework could be beneficial in various real-world applications. 
\section{Related Work}

\textbf{Open-domain question answering.} 
Open-domain question answering (ODQA) \citep{voorhees1999trec} is a task that requires responding to factual questions using external knowledge sources \citep{zhu2015aligning, nagel2016common}.  
Recently, there has been significant research interest in ODQA systems, under a framework known as the \textit{retriever-and-read} system \citep{chen2017reading}. 
The role of \textit{retriever} is to extract the relevant pieces of information from the given knowledge sources. 
For the retriever, there are two different popular methods: one is a lexical-based retriever, \textit{e.g.}, TF-IDF or BM25 \citep{robertson2009probabilistic}, and the other is a sentence embedding-based retriever such as DPR \citep{karpukhin2020dense} or Contriver \citep{izacard2021unsupervised}.
On the other hand, the \textit{reader} is responsible for aggregating and reasoning with the retrieved information to generate answers. 
Usually, recent transformer-based language models (LMs) such as BERT \citep{kenton2019bert} or T5 \citep{raffel2020exploring} are widely adopted for the reader after fine-tuning. 
In contrast, LLMs exhibit comparable performance or outperform in QA without fine-tuning \citep{kamalloo2023evaluating, shi2023replug}, which indicates a potential to serve as a universal QA system \citep{xuan2023evaluation}. 

\textbf{Retrieval-augmented language models.}
Similar to enhancing QA systems with retriever in ODQA, augmenting LMs with relevant information retrieved from external knowledge sources has been demonstrated as an effective way to improve the performance of LMs on various NLP tasks \citep{guu2020retrieval, lazaridou2022internet, min2022nonparametric, liu2023reta}, by reducing hallucination of LLMs and leveraging external knowledge which is not seen during pre-training.
To construct such retrieval-augmented LMs, the standard approach is conducting additional fine-tuning to learn how to incorporate the retrieved information \citep{guu2020retrieval, borgeaud2022improving, izacard2022few}.
However, when considering the recent nature of LLMs with increasing scale and providing black-box API only, such a direction becomes less attractive. 
One promising direction to address this challenge is investigating a better \textit{prompting} \citep{brown2020language}, which incorporates the retrieved information as additional inputs in a sophisticated way.
However, this direction has been only limitedly explored. 
Appending the retrieval \citep{si2023prompting, trivedi2022interleaving} is a common practice for prompting, but \citet{liu2023lost} recently revealed its limitation in utilizing the retrieved information.
Aggregating the predictions from each retrieved passage has been also explored \citep{lazaridou2022internet,shi2023replug}, but LLMs can't see a full context of retrieved information in this case. 
More discussions about the summarization of retrieval in open-domain context are in Appendix \ref{app:related_work}. 
\section{Summarized Retrieval for Question Answering}
\subsection{Overview and problem description}\label{section3.1}

\begin{figure*}[t]
    \centering
    \vspace{-0.1in}
    \includegraphics[width=1.0\textwidth]{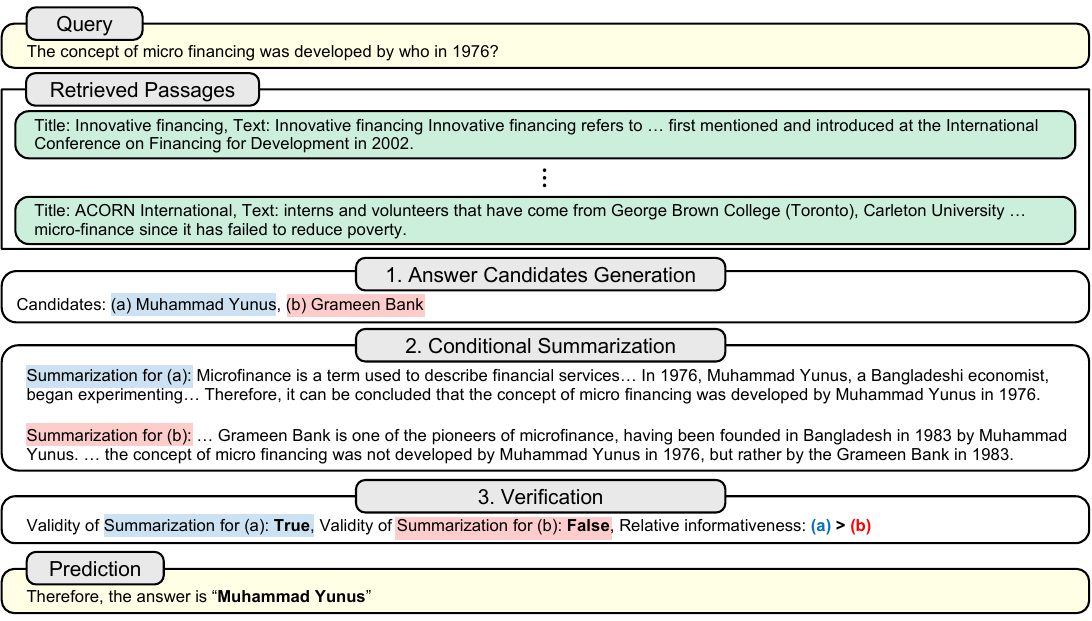}
    \vspace{-0.1in}
    \caption{Example of QA with the proposed \name{} framework. Given a query question and relevant passages retrieved by an external method, \textit{e.g.}, BM25 \citep{robertson2009probabilistic}, a large language model, \textit{e.g.}, ChatGPT, needs to predict the answer. To improve this, \name{} first generates multiple answer candidates via prompting, and then conditionally summarizes the retrieved passages to support each candidate. By comparing the validity and relative informativeness of summaries, \name{} selects the most plausible candidate as a final prediction. 
    }
    \vspace{-0.0in}
    \label{fig:example}
\end{figure*}

\textbf{Overview.} 
In this section, we present our framework, coined Summarized Retrieval (\name{}) to enhance ODQA performance of LLMs, by proposing an improved way to incorporate retrieved passages for the prediction.
Our main idea is to construct multiple summaries of the retrieved passages conditioned with each of a few answer candidates, and predict the most plausible candidate as the answer after evaluating the validity and relative informativeness of summaries.
In Sections \ref{section3.2} and \ref{section3.3}, we present the details to generate the summarizations and evaluate them. 
Figure \ref{fig:example} presents the specific example of QA procedure via \name{}.

\textbf{Problem description.} 
Open-domain question answering (ODQA) is an extension of QA tasks that answer questions that require background knowledge by leveraging an external database. 
In order to answer the given question $q$, the ODQA system typically follows \textit{retrieve-and-read} framework \citep{chen2017reading, lee2019latent}, where the \textit{retriever} finds the informative passages $C^{+}_{N}$ from the whole corpus $C$, and the \textit{reader} exploits the retrieved passages to decide the answer $a$, which can be formulated as follows: 
\begin{equation}\label{eq:odqa_form}
 C^{+}_{N} = \texttt{Retriever}(q, C, N) \;\; \text{and} \;\; \widehat{a} = \texttt{Reader}(q, C^{+}_{N}),
\end{equation}
where $N$ is the number of retrieved passages and $\widehat{a}$ is the predicted answer.

In this work, we focus on improving a prompting method for an LLM-based ODQA system.
Specifically, we adopt the existing \textit{retriever} method, \textit{e.g.}, BM25 \citep{robertson2009probabilistic} or Contriever \citep{izacard2021unsupervised}, with the dataset-specific corpus. 
For the \textit{reader} method, we use LLMs, denoted by $\mathcal{M}$, such as ChatGPT \citep{sun2023chatgpt} or LLaMA-2 \citep{touvron2023llama}, by incorporating the retrieved passages via \textit{prompting} \citep{brown2020language} without additional training.   
For example, with a prompt $p(q, C^{+}_{N})=\text{``}\texttt{Reading passages } C^{+}_{N}, \texttt{ answer to question } q$'', the prediction $\widehat{a}$ is obtained from $\mathcal{M}$, \textit{i.e.},  $\widehat{a}=\mathcal{M}\left(p(q, C^{+}_{N})\right)$.

\subsection{Conditional summarization of retrieved passages}\label{section3.2}

To better exploit the retrieved passages with LLMs, \name{} first summarizes them conditioned on each of a few potential answer candidates. 
This \textit{conditional} summarization of retrieved passages would include the specific contexts supporting a given answer candidate, compared to the generic summarization focusing on the wide coverage for the retrieved passages.
Specifically, \name{} first generates answer candidates and then conducts conditional summarization.

\textbf{Candidates generation.} 
Given a question $q$, retrieved passages $C^{+}_{N}$, and LLM $\mathcal{M}$, we first generate $K$ answer candidates $\widetilde{\mathbf{y}}=[\widetilde{y}_{1}, \dots, \widetilde{y}_{K}]$ using a prompt $p_{\tt can}$ designed for candidate generation from $q$ and $C^{+}_{N}$:
\begin{equation}\label{eq.2}
    \widetilde{\mathbf{y}} = \mathcal{M}\left(p_{\tt can}(q, C^{+}_{N})\right).
\end{equation}
In Figure \ref{fig:example}, one can observe the example of generated candidates.
It is noticeable that the previous works utilized stochastic decoding to generate multiple answer candidates \citep{lazaridou2022internet, weng2022large}. 
However, we empirically observe that explicitly prompting an LLM to generate $K$ potential candidates outputs more diverse and high-quality candidates.

\textbf{Candidate-conditioned summarization.} 
Next, we \textit{conditionally} summarize the retrieved passages $C^{+}_{N}$ focusing on including the relevant contexts to validate each candidate $\widetilde{y}_{k} \in \widetilde{\mathbf{y}}$ as an answer to $q$:
\begin{equation}\label{eq.3}
    s_{k} = \mathcal{M}\left(p_{\tt sum}(q, C^{+}_{N}, y_{k})\right)\; \text{for}\;~ k=1,\dots,K
\end{equation}
\begin{wrapfigure}[11]{r}{0.3\textwidth}
	\vspace{-0.5cm}
	{
	\hspace{0.0cm}
	\includegraphics[width=40mm]{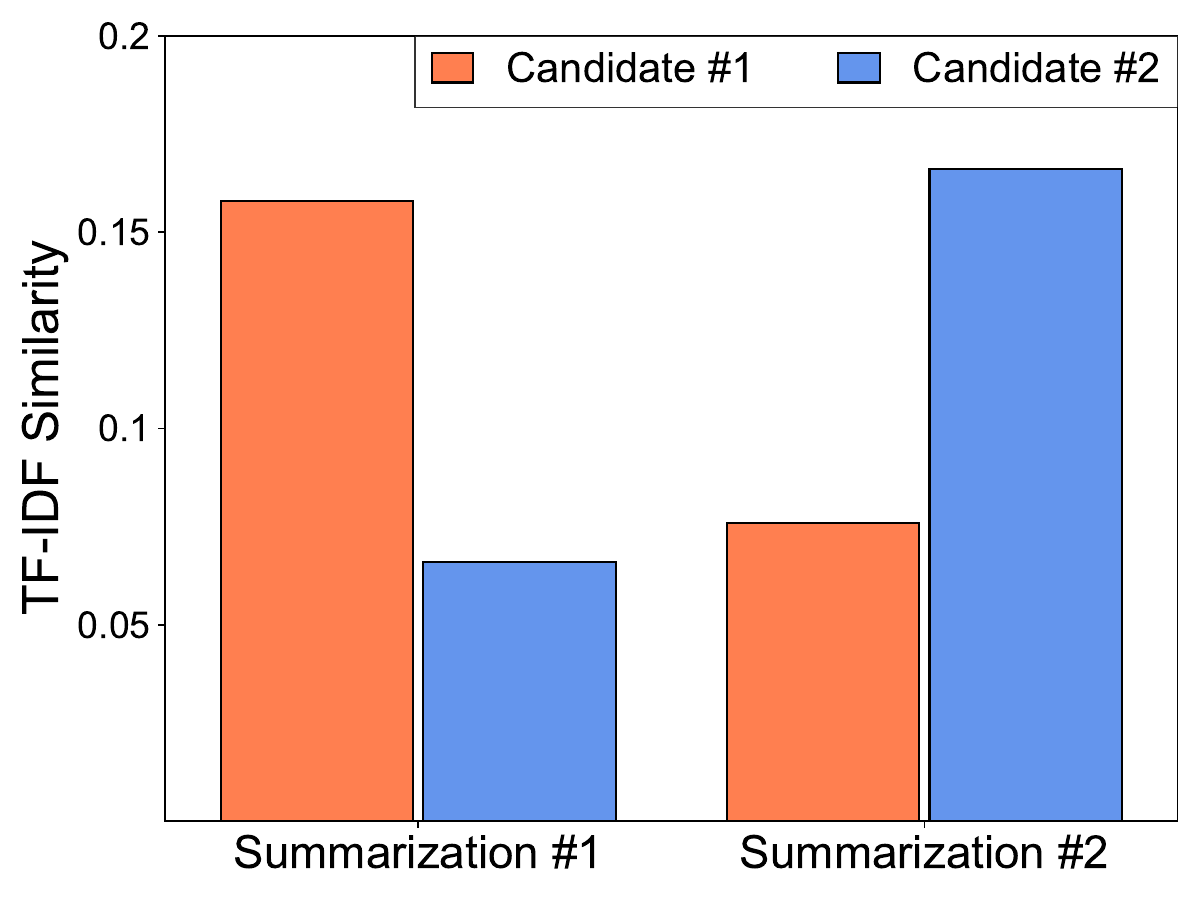}
	\vspace{-0.4cm}
	\caption{TF-IDF overlap between candidates and conditional summarizations.}\label{tfidf}
	}
\end{wrapfigure}
where $p_{\tt sum}$ is a prompt to obtain the conditional summarization $s_{k}$ from $q$, $C^{+}_{N}$, and $\widetilde{y}_{k}$.
We present some examples of the generated summarizations in Figure \ref{fig:example}, and more examples are in Appendix~\ref{appendixC}. 
Remarkably, the generated summarizations effectively reduce the given passages by focusing on extracting the candidate-relevant contexts (\textit{e.g.}, 1035 words of retrieved passages $\rightarrow$ 93 words of summarization). 
Also, we verify that the contexts of the generated summarization are specialized on a given answer candidate; when we measure TF-IDF \citep{chowdhury2010introduction} based text similarity between two candidates and two conditional summarizations from each candidate (\textit{e.g.}, summarization \#1 is generated to support answer candidate \#1) on Natural Question dataset \citep{kwiatkowski2019natural} in Figure \ref{tfidf}, the summarization exhibits a higher similarity with the corresponding candidate than the other candidate. 

\subsection{Selective prediction via verification of summarizations}\label{section3.3}

Then, using the generated summarizations, \name{} confirms the most plausible answer among the candidate set for the prediction.
Our key intuition is that the quality (\textit{e.g.}, factuality, logicality, and readability) of the generated summarizations would vary depending on the plausibility of answer candidates, so as more plausible the answer, the corresponding summarization also will be more plausible. 
Then, LLMs can find the most plausible summarization among these multiple summarizations if a proper evaluation way is given. 
To this end, we propose to evaluate the generated summarizations with \textit{instance-wise} validity and \textit{pair-wise} ranking among them. 

\textbf{Instance-wise validity.} 
First, we evaluate the validity of each summarization $s_{k}$ whether it is not a degenerated case as the provided passages are not enough to support $\widetilde{y}_{k}$, or it properly supports the given answer candidate $\widetilde{y}_{k}$, rather than the other candidate $\widetilde{y}_{i}, ~ i \neq k$.\footnote{We present such failure cases in Appendix \ref{appendixD}.}  
To be specific, we measure a validity $v_{k}$ of each summarization $s_{k}$ using a prompt $p_{\tt val}$ designed for the validation:
\begin{equation}\label{eq.4}
    v(s_{k}) = 1, ~\text{when}~ \mathcal{M}\left(p_{\tt val}(q, y_{k}, s_{k})\right) = \text{True} ~~~\text{or}~~~ v(s_{k}) = 0, ~ \text{else}.
\end{equation}

\textbf{Pair-wise ranking.} 
In addition, we evaluate how the given summarization $s_{k}$ is \textit{relatively informative} to answer the question $q$, among all summaries $S_{K}=\{s_{k}\}_{k=1}^{K}$.
To this end, we measure a ranking $r_{k}$ using a pair-wise ranking prompts \citep{qin2023large, sun2023chatgpt}: 
\begin{equation}\label{eq.5}
    r(s_{k},S_{K}) = \sum_{i \neq k}^{K} r_{\tt pair}(s_{k},s_{i}), ~ r_{\tt pair}(s_{k},s_{i}) = \begin{cases} 1, & \mathcal{M}\left(p_{\tt rank}(q, s_{k}, s_{i})\right) = s_{k} \\ 0, & \mathcal{M}\left(p_{\tt rank}(q, s_{k}, s_{i})\right) = s_{i} \\ 0.5, & \text{else} \end{cases}, 
\end{equation}
where $p_{\tt rank}$ is a prompt to determine which is relatively more informative one to answer the question by comparing two summaries. 
To prevent the order bias of LLMs \citep{zhao2021calibrate}, we query the same pair of summaries twice by changing their order at the prompt $p_{\tt rank}$.

Finally, \name{} makes a final prediction $\widehat{a}$ by incorporating both $v(s_{k})$ and $r(s_{k}, S_{K})$:
\begin{equation}\label{eq.6}
    \widehat{a} = \widetilde{y}_{k^{*}}, \; k^{*} = \argmax_{k} v(s_{k}) + r(s_{k}, S_{K}), 
\end{equation}
\textit{i.e.}, both validity and ranking scores are \textit{equally} contributed.
Algorithm \ref{alg:ours} summarizes the formal procedure of \name{}.
We also highlight that the common prompts are shared across different datasets and LLMs, and the used prompts $p_{\tt can}, p_{\tt sum}, p_{\tt val}, p_{\tt rank}$ are presented in Appendix \ref{appendixB}.

\begin{algorithm}[t]
    \caption{\texttt{\name{}} algorithm}\label{alg:ours}
    \begin{algorithmic}[1]
        \State \textbf{Input:} Large language model $\mathcal{M}$, question $q$, $N$ retrieved passages $C^{+}_{N}$, candidate number $K$
        \State \textbf{Answer Candidate Generation:} $\widetilde{\mathbf{y}} = \mathcal{M}\left(p_{\tt can}(q, C^{+}_{N})\right), ~ \widetilde{\mathbf{y}}=[\widetilde{y}_{1}, \dots, \widetilde{y}_{K}]$
        \State \textbf{Conditional Summarization:} $s_{k} = \mathcal{M}\left(p_{\tt sum}(q, C^{+}_{N}, y_{k})\right)\; \text{for} \;~ k=1,\dots,K$
        \State \textcolor{black}{\textbf{Instance-wise Validation:}} $v(s_{k}) \leftarrow$ Eq.~\ref{eq.4} with $\mathcal{M}\left(p_{\tt val}(q, s_{k})\right)$
        \State \textcolor{black}{\textbf{Pair-wise Ranking:}} $r(s_{k},S_{K}),~ r_{\tt pair}(s_{k},s_{i}) \leftarrow$ Eq.~\ref{eq.5} with $\mathcal{M}\left(p_{\tt rank}(q, s_{k}, s_{i})\right)$
        \State \textbf{Output:} Prediction $\widehat{a} = \widetilde{y}_{k^{*}}, ~ k^{*} = \argmax_{k} v(s_{k}) + r(s_{k}, S_{K})$   
    \end{algorithmic}
\end{algorithm}

\section{Experiments}\label{sec:4}

In this section, we design our experiments to investigate the following questions:
\begin{itemize}[leftmargin=5.5mm,topsep=0pt]
    \item[$\circ$] Does \name{} improve the accuracy of LLMs on various ODQA datasets? (Table~\ref{table:main_odqa}) \vspace{-0.04in}
    \item[$\circ$] Is \name{} generalizable across various retrieval methods and LLMs? (Table~\ref{table:diff_retrieval_llm_em}) \vspace{-0.04in}
    \item[$\circ$] What is the effect of each component in \name{}? (Table~\ref{table:ablation}) \vspace{-0.04in}
    \item[$\circ$] Is \name{}'s summarization a good rationale for the answer? (Table~\ref{table:rerank} \& Figure~\ref{fig:fig3_analysis}) \vspace{-0.04in}
\end{itemize}

\subsection{Setups}\label{sec:4.1}

\textbf{Evaluation datasets.}
For all experiments, we measure zero-shot QA accuracy with the four different ODQA datasets:
(1) Natural Questions (NQ) \citep{kwiatkowski2019natural},
(2) WebQuestions (WebQ) \citep{berant2013semantic}, 
(3) 2WikiMulti-hopQA (2Wiki) \citep{ho2020constructing}, and (4) HotpotQA \citep{yang2018hotpotqa}. 
For NQ and WebQ, we use their original test splits and 21M English Wikipedia dump \citep{karpukhin2020dense} as the source passages for the retrieval.
For 2Wiki and HotpotQA, we use the subsampled splits released by \cite{trivedi2022interleaving}, along with the corresponding corpus for each data.
For the experiments with LLaMA2-chat (Table \ref{table:diff_retrieval_llm_em}) and more analyses (Section \ref{sec:4.3}), we took 500 randomly subsampled examples of NQ and WebQ datasets for efficient experiments considering limited computing resources, and denoted these datasets NQ$^{*}$ and WebQ$^{*}$, respectively.
As evaluation metrics,  we calculate the exact match (EM) and F1 score. 
The EM accuracy is the ratio of correct answers in the test dataset, where a given prediction is considered correct if it coincides with one of the gold answers.
The F1 score measures the overlap between bags of tokens in the prediction and the gold answer.
We normalize the predictions and answers (\textit{i.e.}, case-folded, and punctuation) to compute the metrics, following the implementation of \cite{rajpurkar2016squad}.

\begin{table*}[t]
\vspace{-0.1in}
\centering
\caption{EM / F1 for different QA methods with ChatGPT on four QA datasets. $N=10$ most relevant passages are retrieved using 
BM25, except \textit{no retrieval}. The best and second best scores are highlighted in \textbf{bold} and \underline{underline}, respectively.
\vspace{0.1in}
}
\scalebox{0.95}{
{
\begin{tabular}{r|cccc|c}
\toprule
Methods / Datasets & NQ & WebQ & 2Wiki & HotpotQA & Average \\
\midrule
No retrieval & 27.6 / \underline{39.0}  & \underline{25.0} / \textbf{38.8} & 21.4 / 24.8 & 22.2 / 31.9 & 24.1 / 33.6 \\ \midrule
Base & \underline{28.4} / 38.8 & 19.6 / 32.5 & \underline{27.4} / \underline{32.8} & 30.8 / 40.3 & \underline{26.6} / \underline{36.1} \\
Rerank & 24.8 / 33.9 & 18.8 / 30.6 & 23.0 / 28.4 & 27.8 / 37.4 & 23.6 / 32.6 \\
RePlug & 26.0 / 35.3 & 18.8 / 31.5 & 23.6 / 28.5 & 28.0 / 37.9 & 24.1 / 33.3 \\
Selection-inference & 24.3 / 32.8 & 17.3 / 28.6 & 22.6 / 29.5 & 30.8 / 39.6 & 23.8 / 32.6 \\
Chain-of-thoughts 
& 22.3 / 31.4 & 15.2 / 27.8 & 19.6 / 22.5 & 25.6 / 31.8 & 20.7 / 28.4 \\ 
Self-verification & 25.2 / 35.4 & 16.1 / 28.5 & 23.2 / 30.5 & \underline{31.6} / \underline{41.8} & 24.0 / 34.1 \\ 
\midrule
\name{} (Ours) & \textbf{33.5} / \textbf{42.3} & \textbf{25.1} / \underline{36.6} & \textbf{32.8} / \textbf{38.1} & \textbf{33.2} / \textbf{43.4}  & \textbf{31.2} / \textbf{40.1} \\
\bottomrule
\end{tabular}}
}
\label{table:main_odqa}
\vspace{-0.1in}
\end{table*}

\textbf{Baselines.} 
We compare \name{} with the following baselines. 
(1) \textit{No retrieval} answers the question with LLMs without the retrieved passages (\textit{i.e.}, closed-book setup).
(2) \textit{Base} appends the retrieved passages as additional inputs of LLMs via prompting.
(3) Line of works for better exploitation of retrieved passages with LLMs: 
\textit{Rerank} \citep{lazaridou2022internet} and \textit{RePlug} adopt an ensemble strategy that makes predictions based on each passage and then aggregates them with specific voting methods. 
Specifically, \textit{Rerank} and \textit{RePlug} utilize TF-IDF and sentence embedding from Contriever, respectively. 
(4) Adapt the works that incorporate intermediate reasoning steps for improved reasoning with LLMs, as summarizing could be viewed as a specific type of reasoning: 
\textit{Selection-inference} \citep{creswell2023selection} measures the ranking of the passages, and conducts interactive answering by adding the passages one by one starting from higher ranked ones.
\textit{Chain-of-thoughts} \citep{kojima2022large}: we add zero-shot Chain-of-thoughts prompting \citep{wei2022chain} into the prompt of \textit{Base}. \textit{Self-verification} \citep{weng2022large} generates answer candidates based on random sampling, then selects the most plausible one by verifying its reasoning with the question from conditional masking.

\textbf{Implementation details.} For the experiments, we use three recent state-of-the-art LLMs: ChatGPT (\texttt{gpt-3.5-turbo-0301}) \citep{openai2022chatgpt}, GPT-4 (\texttt{gpt-4-0613}) \citep{openai2023gpt4}, and LLaMA2-chat-70B \citep{touvron2023llama}. 
We use a temperature of $0.0$ when calling the API or greedy decoding for LLaMA, to remove the effect of random sampling \citep{sun2023chatgpt}. 
For the retrieval methods, we use three different approaches: BM25 \citep{robertson2009probabilistic}, DPR-multi (DPR) \citep{karpukhin2020dense}, and Contriever \citep{izacard2021unsupervised}. 
We use the implementations in Elasticsearch for BM25, and BEIR for DPR and Contriever, respectively.\footnote{\url{https://www.elastic.co/}, \url{https://github.com/beir-cellar/beir}}
In the case of \name{}, we use the same prompts across the different datasets, and they are presented in Appendix \ref{appendixB}.
Also, we use a fixed value of $K=2$ during the experiments since we observe that the improvements by increasing $K$ are limited, as shown in Appendix \ref{appendixC}. 
When there are multiple candidates with equal plausibility (Eq. \ref{eq.6}), then \name{} selects the one generated earlier in Eq. \ref{eq.2}.  

\subsection{Main results}\label{sec:4.2}

\begin{table*}[t]
\vspace{-0.1in}
\centering
\caption{EM with different configurations of LLMs and retrieval methods on four QA datasets. $N=10$ most relevant passages are commonly retrieved. F1 scores are reported in Table \ref{table:diff_retrieval_llm_f1}. For LLaMA2-chat, we conducted experiments on NQ$^{*}$ and WebQ$^{*}$ and the results are indicated by $^{*}$.
}
\vspace{0.1in}
\scalebox{0.8}{
{
\begin{tabular}{r|cccccc|cc|cc}
\toprule
& \multicolumn{6}{c|}{ChatGPT} & \multicolumn{2}{c|}{GPT-4} & \multicolumn{2}{c}{LLaMA2-chat} \\ \cmidrule(l){2-7} \cmidrule(l){8-11}
Datasets & BM25 & + \name{} & DPR & + \name{} & Contriever & + \name{} & BM25 & + \name{} & BM25 & + \name{} \\ \midrule
NQ & 28.4 & \textbf{33.5}  & 36.1 & \textbf{41.0} & 35.8 & \textbf{40.4} & 30.2 & \textbf{32.4}  & 18.6$^{*}$ & \textbf{30.4}$^{*}$ \\
WebQ & 19.6 & \textbf{25.1} & 23.2 & \textbf{27.3} & 22.5 & \textbf{28.7} & 21.5 & \textbf{21.7}  & 16.0$^{*}$ & \textbf{24.0}$^{*}$ \\ 
2Wiki & 27.4 & \textbf{32.8} & 19.2 & \textbf{21.4} & 27.2 & \textbf{32.6} & 34.8 & \textbf{38.2}  & 20.2 & \textbf{27.8} \\ 
HotpotQA & 30.8 & \textbf{33.2} & 25.6 & \textbf{27.4} & 32.2 & \textbf{33.6} & 34.8 & \textbf{40.6}  & 24.0 & \textbf{28.0} \\ 
\midrule
Average & 26.6 & \textbf{31.2} & 26.0 & \textbf{29.3} & 29.4 & \textbf{33.8} & 30.3 & \textbf{33.2} & 19.7 & \textbf{27.6} \\ 
\bottomrule
\end{tabular}}
}
\label{table:diff_retrieval_llm_em}
\vspace{-0.15in}
\end{table*}

Table \ref{table:main_odqa} summarizes the experimental results on four different ODQA datasets, under ChatGPT with $N=10$ retrieved passages using BM25. 
First, augmenting the retrieved passages with prompting is effective in improving ODQA accuracies of LLMs. 
For example, the average EM across four ODQA datasets is increased from $24.1$ to $26.6$. 
Somewhat surprisingly, we observe that \textit{Base} outperforms other sophisticated baselines overall; this inefficiency of previous methods might be a result of a more challenging yet practical experimental setup. 
For example, we assume the zero-shot QA rather than few-shot setups, and also consider general black-box APIs for LLMs which do not provide the output probability. 
In contrast, one can observe that \name{} successfully improves QA accuracy of LLMs by effectively exploiting the retrieved passages.
In particular, \name{} exhibits 4.6\%/4.0\% absolute EM/F1 improvements in the average, compared to na\"ively appending the retrieved passages. 

We further demonstrate the compatibility of \name{} across various LLMs and retrieval methods. 
Specifically, in addition to ChatGPT and BM25 considered in Table \ref{table:main_odqa}, we run experiments on three different LLMs (GPT-4, and LLaMA2-chat) and two different retrieval methods (DPR and Contriever).
In Table \ref{table:diff_retrieval_llm_em}, we compare EM metric of \name{} with the baseline that simply appends the retrieved passages. 
Here, ODQA performance significantly depends on the retrieval methods and types of LLMs; for example, using Contriever instead of BM25 makes 2.8\% average EM improvements, and using GPT-4 instead of ChatGPT makes 3.7\% average EM improvements, respectively.
Overall, one can observe that \name{} consistently improves ODQA accuracy regardless of types of LLMs and retrieval methods, with 4.6\% average EM improvements. 
More interestingly, \name{} successfully improves average EM scores of LLaMA2-chat as 7.9\%, a state-of-the-art open-sourced LLM, which further indicates the practical usefulness of \name{} as a simple yet effective solution for ODQA for the open source research community.
The F1 results are presented in Appendix~\ref{appx:more_config}.

\subsection{Additional analyses}\label{sec:4.3}

\begin{table*}[t]
\centering
\caption{Ablation and more analyses. EM / F1 with ChatGPT are compared on four QA datasets. $N=10$ most relevant passages are retrieved using BM25. The best scores are highlighted in \textbf{bold}.
\vspace{0.1in}
}
\scalebox{0.9}{
{
\begin{tabular}{l|cccc|c}
\toprule
Methods / Datasets & NQ$^{*}$ & WebQ$^{*}$ & 2Wiki & HotpotQA & Average \\ \midrule
Base & 29.4 / 41.7 & 19.4 / 32.2 
& 27.4 / 32.8 & 30.8 / 40.3 
& 26.8 / 36.8 \\ 
Conditional summarizations & 30.4 / 40.9 & 20.8 / 33.5 & 29.2 / 34.5 & 33.0 / \textbf{43.4} & 28.4 / 38.1  \\
+ Pair-wise ranking & 30.6 / 41.2 & 21.6 / 34.8
& 31.0 / 36.0 & 30.6 / 40.7
& 28.5 / 38.2  \\
+ Instance-wise validity (\name{})  & \textbf{35.6} / {44.9} & \textbf{23.2} / \textbf{36.5} 
& \textbf{32.8} / \textbf{38.1} & {33.2} / \textbf{{43.4}}
& \textbf{31.2} / \textbf{40.7} \\
\midrule
MCQ prompt & {35.2} / \textbf{45.3} & 22.4 / 35.1 
& 30.4 / 36.1 & 31.0 / 41.5  
& 29.8 / 39.5 \\ 
Sum-and-pred (Gen) & 26.4 / 37.8 & 19.8 / 32.6 
& 25.6 / 32.3 & \textbf{{33.8}} / {43.3}
& 27.3 / 37.1  \\ 
\bottomrule
\end{tabular}}
}
\label{table:ablation}
\vspace{-0.1in}
\end{table*}

\begin{table*}[t]
\vspace{-0.15in}
\centering
\caption{
Comparison as reranking method. EM / F1 with ChatGPT are compared on four QA datasets. A single most relevant passage is selected among $N=10$ passages retrieved by BM25. The best scores are highlighted in \textbf{bold}.
\vspace{0.1in}
}
\scalebox{0.95}{
{
\begin{tabular}{r|cccc|c}
\toprule
Datasets / Methods & NQ$^{*}$ & WebQ$^{*}$ & 2Wiki & HotpotQA & Average \\ \midrule
BM25 & 12.6 / 18.8 & 9.0 / 17.6 
& 14.8 / 18.1 & 21.8 / 28.3 
& 14.6 / 20.7 \\
Sent-encoder (q) & 18.2 / 26.4 & 11.4 / 21.1
& 14.8 / 18.1 & 20.2 / 27.2
& 16.2 / 23.2  \\
LLM-rerank & 20.0 / 28.4 & 14.2 / 24.4 
& \textbf{18.2} / 21.3 & 26.0 / 34.4  
& 19.6 / 27.1 \\ 
Sent-encoder (Gen) & 21.2 / 31.7 & 13.6 / 25.3 
& 17.8 / 21.1 & 27.0 / 34.3
& 19.9 / 28.1  \\ \midrule
Sent-encoder (\name{}) & \textbf{23.2} / \textbf{32.5} & \textbf{15.4} / \textbf{28.0} 
& 18.0 / \textbf{21.5} & \textbf{28.8} / \textbf{36.7}
& \textbf{21.4} / \textbf{29.7} \\ 
\bottomrule
\end{tabular}}
}
\label{table:rerank}
\end{table*}

\begin{figure*}[t]
\vspace{-0.1in}
\begin{center}
    {
    \subfigure[Different number of retrieval]
        {
        \includegraphics[width=0.31\textwidth]{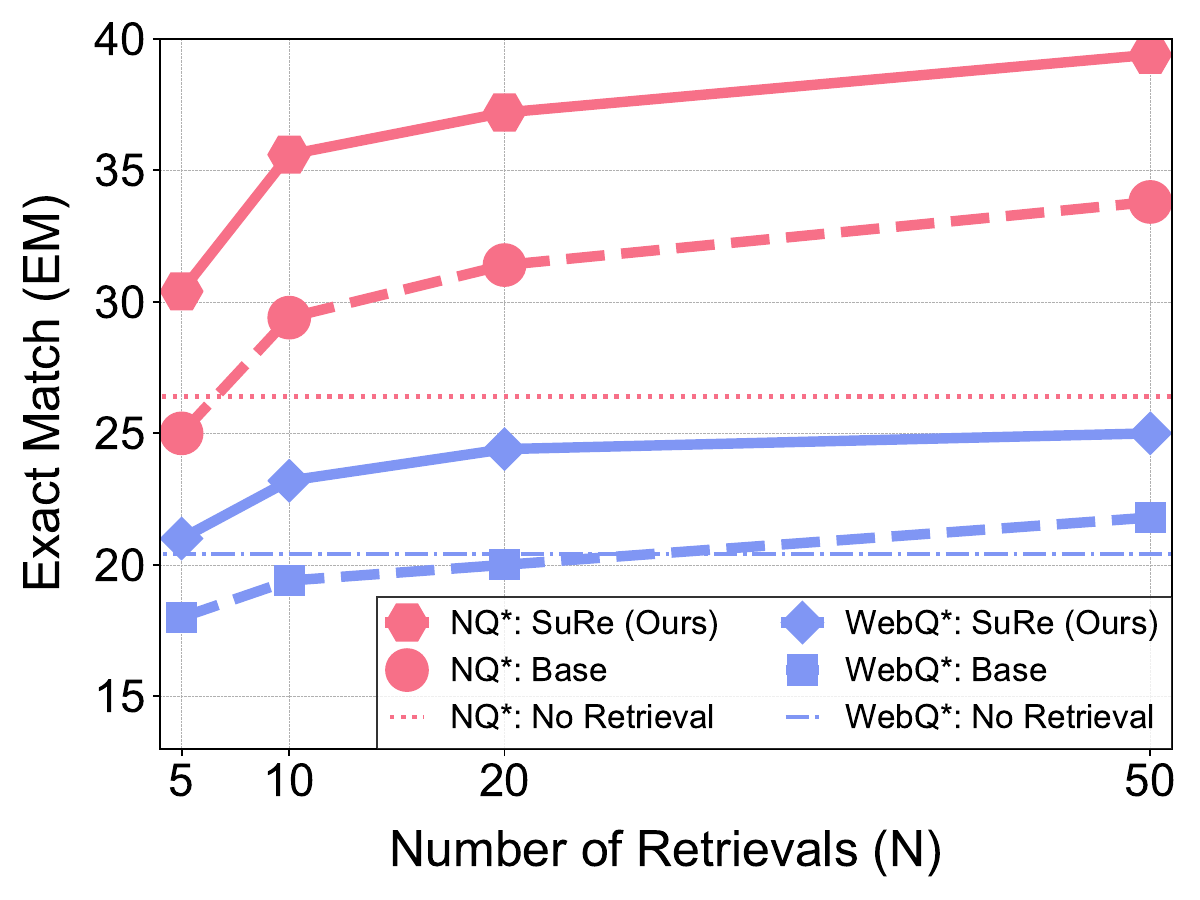}
        \label{fig:fig3a}
        }
    \subfigure[GPT-4 evaluation]
        {
        \includegraphics[width=0.31\textwidth]{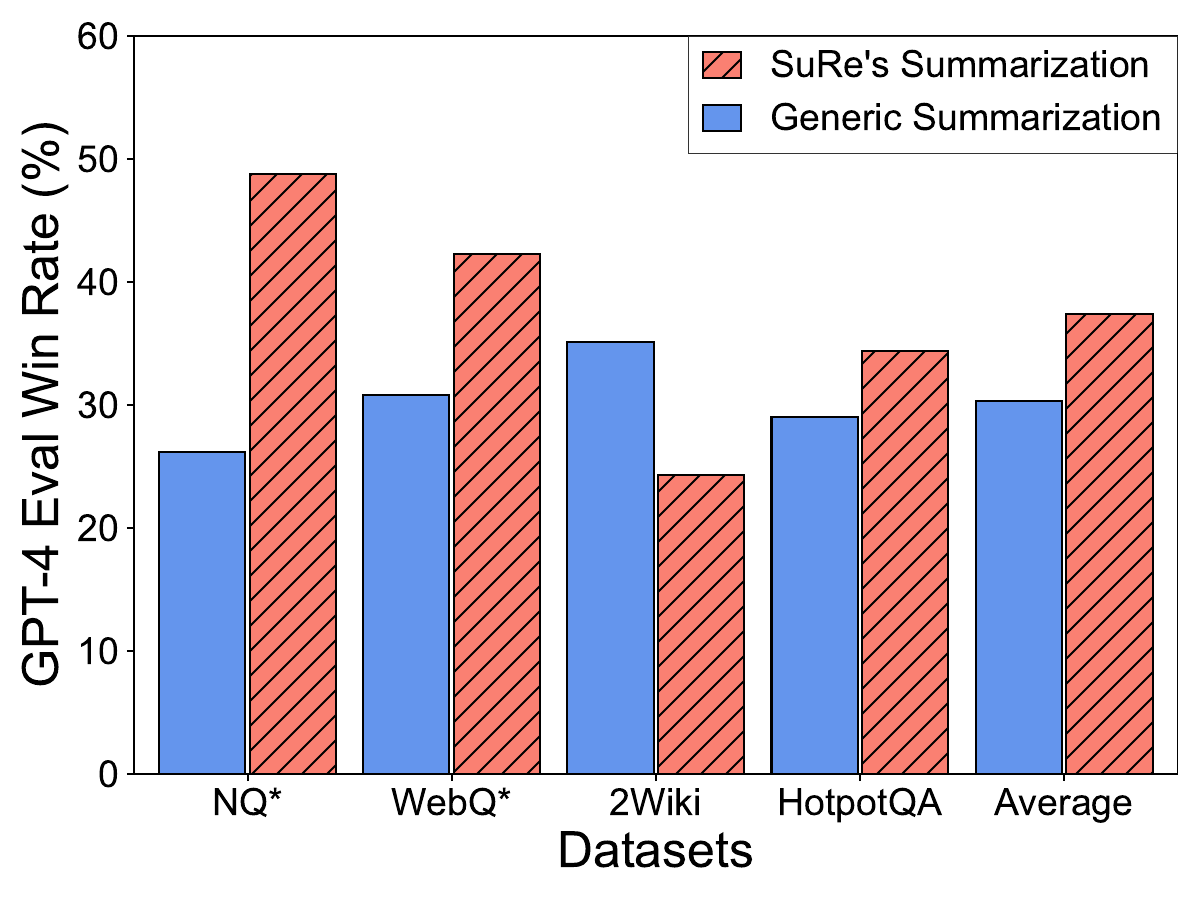}
        \label{fig:fig3b}
        } 
    \subfigure[Human preference]
        {
        \includegraphics[width=0.31\textwidth]{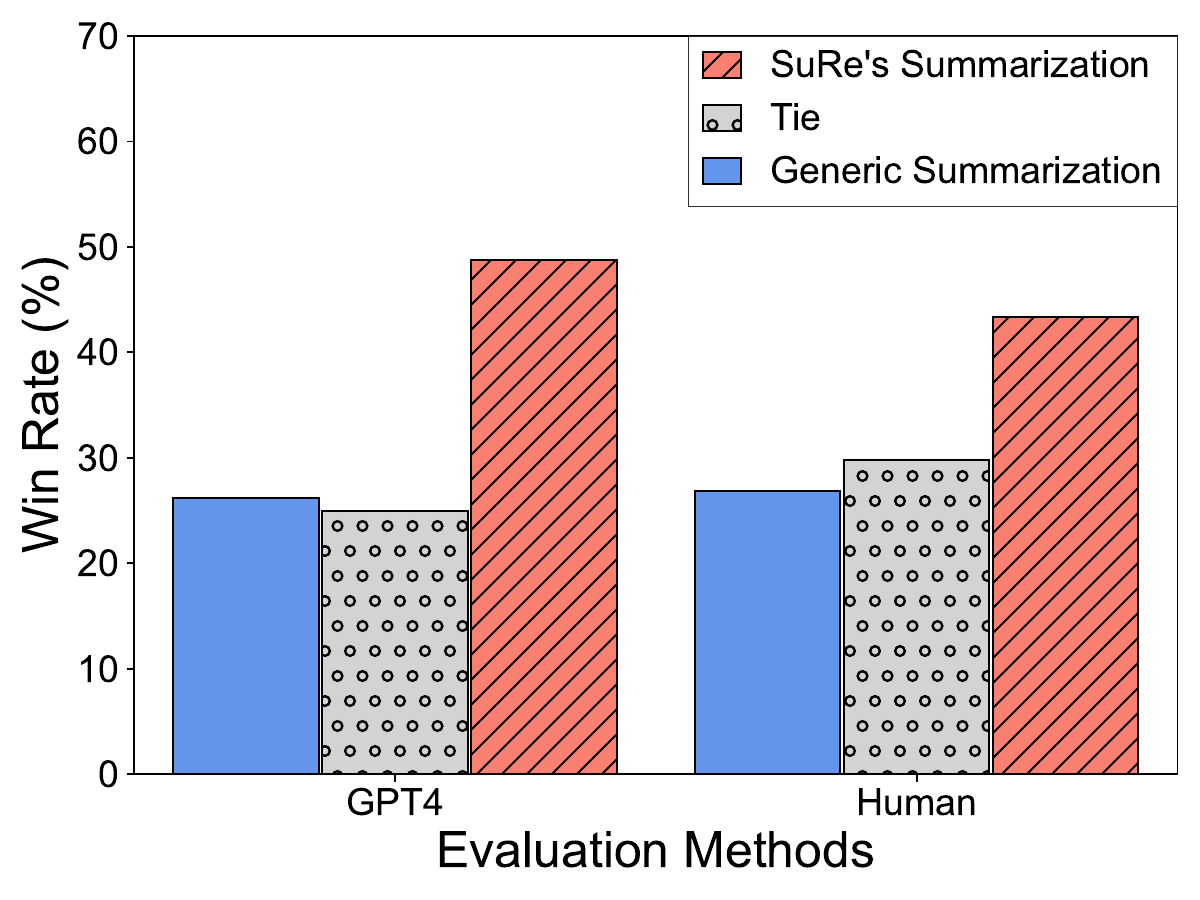}
        \label{fig:fig3c}
        }
    }
\end{center}
\vspace{-0.1in}
\caption{(a) EM with different numbers of retrieved passages ($N$) under ChatGPT and BM25. (b) Comparison between \name{}'s summarization and generic summarization via GPT-4 evaluation \citep{liu2023gpteval}. (c) Human preference between \name{}'s summarization and generic summarization on 84 samples of NQ$^{*}$, along with GPT-4 evaluation. More results are in Appendix \ref{appendixE}.}
\vspace{-0.15in}
\label{fig:fig3_analysis}
\end{figure*}

In this section, we conduct additional analyses of \name{}. 
We conduct experiments using ChatGPT as an LLM, BM25 as a retriever, NQ$^{*}$ and WebQ$^{*}$ as datasets.

\textbf{Ablation and more analysis of \name{}}.
First, we compare the following methods for the ablation of \name{}: (1) \textit{Base}: appends the retrieved passages to inputs, (2) \textit{+ Conditional summarizations}: additionally appends all the conditional summarizations,  
(3) \textit{+ Pair-wise ranking}: selects the summarization with only ranking (Eq.~\ref{eq.4}), and (4) \textit{+ Instance-wise validity}: selects the summarization with both ranking and validity, \textit{i.e.}, \name{}. 
In addition, we consider two different methods to further analyze where the effectiveness of \name{} comes from.
(5) \textit{MCQ prompt}: composes Multiple Choice Questions by
generating the answer candidates via prompting (Eq.~\ref{eq.2}) and using them as possible choices for prediction by appending them to input prompt \citep{robinson2023leveraging} 
 (more details in Appendix \ref{A.7}), (6) \textit{Sum-and-pred (Gen)}: instead of \textit{conditional} summarization, it generates generic summarization and predicts the answer based on it. 
We present the results in Table \ref{table:ablation}. 

First, constructing conditional summarizations improves performance as they can extract specialized contexts for a given question and its answer candidates.
Next, incorporating the evaluation on the instance-wise validity of each summarization significantly improves the performance compared to only considering the ranking among summarizations, as it enables more precise selection by adding the assessment regarding the relevance and coherence of the summarization in relation to the given question and prediction pair. 
Also, a simple aggregation of generated answer candidates in the prompt shows improvement, which indicates the effectiveness of our generated candidates. 
However, this method becomes inefficient when the given question requires more complex reasoning to answer.     
Lastly, using generic summarization is effective in improving ODQA with LLMs by providing concentrated and brief contexts and addressing the difficulty from the long context \citep{liu2023lost}. 
However, the gain is significantly limited compared to \name{}, which demonstrates that the key components of \name{} are conditional summarization and comparison, rather than simply providing compressed contexts.  

\textbf{Different number of retrieval.}
Next, we investigate the effect of the number of retrieved passages ($N$).
Increasing $N$ is one of the most intuitive ways to improve the performance of \textit{retrieve-and-read} system by providing more extensive information \citep{karpukhin2020dense}, and hence it is natural to expect that similar positive results could be observed with retrieval-augmented LLMs. 
However, on the other hand, its effectiveness could be limited as LLMs could fail to handle long input contexts \citep{liu2023lost}.
To verify the effect of different $N$ on retrieval-augmented LLMs using prompting, we measure EM of ChatGPT and BM25 with varied $N$.
In Figure \ref{fig:fig3a}, we present the results of \textit{Base} and \name{} on NQ$^{*}$ and WebQ$^{*}$. 
First, we observe that the accuracy of retrieval-augmented LLMs significantly depends on $N$; when a small number of retrieved passages is only available, the performance of \textit{Base} could be even behind the performance without retrieval, as it restricts the prediction within the limited contexts. 
As $N$ increases, its performance is increased and takes benefit from the retrieval system. 
With \name{}, the accuracy of LLMs could be improved even with the small number of retrievals ($N=5$), and it achieves better accuracy with larger $N$. 

\textbf{Effectiveness for finding important passages.}
In previous experiments, we mainly focus on demonstrating the effectiveness of \name{} for improving QA accuracy.
While the accurate answer is the most important feature of the QA system, providing the proper rationale for the answer is another important feature, especially in LLM-based systems for reliable usage by users such as search engines.  
One of the standard approaches for this is explicitly enumerating the most relevant retrieved passages based on the specific scoring method, which is often called \textit{Re-ranking} \citep{nguyen2016ms, izacard2022few}. 
To explore the advantages of \name{} in this aspect, we measure QA accuracy of ChatGPT augmented with the one passage considered to be most relevant with a specific reranking method within $N=10$ originally retrieved passages with BM25.
To extract such a reranking method for \name{}, we use the cosine similarity between the sentence embeddings \citep{reimers2019sentence} of the generated summarization and the retrieved passages, denoted by \textit{Sent-encoder (\name{})}.
Then, we compare it with the following baselines for reranking: (1) \textit{BM25}: original retrieval score, \textit{i.e.}, no reranking, (2) \textit{Sent-encoder (q)}: sentence-encoder-based reranking using the similarity between retrieved passages and question \citep{nguyen2016ms}, (3) \textit{LLM-rerank}: LLM-based reranking \citep{sun2023chatgpt}, and (4) \textit{Sent-encoder (Gen)}: sentence-encoder-based reranking using the similarity between retrieved passages and generic summarization. 
The results are presented in Table \ref{table:rerank}. Here, we observe that all the reranking methods are effective compared to no reranking. In addition, LLM-based reranking shows a higher accuracy, while \name{}'s similarity-based reranking outperforms all the baselines, demonstrating the superiority of \name{}.

\begin{figure*}[t]
\vspace{-0.2in}
\begin{center}
    \includegraphics[width=1.0\textwidth]{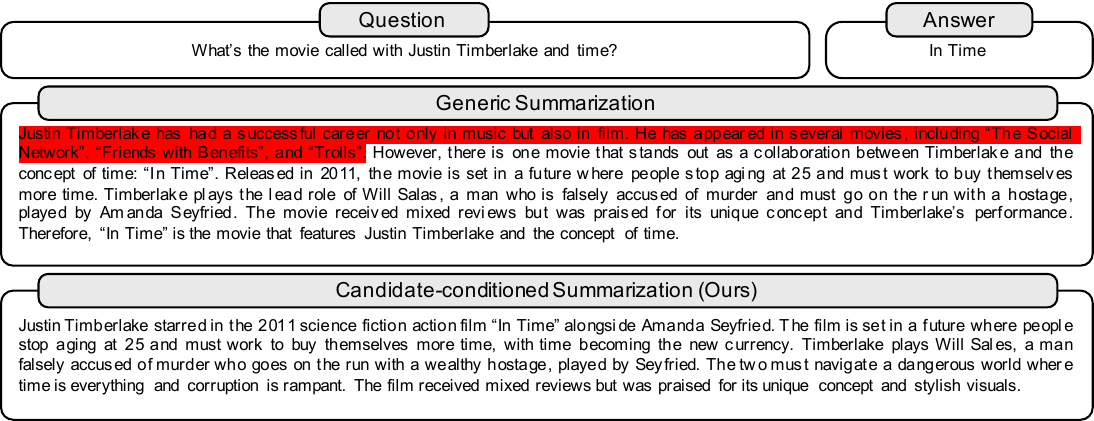}
\end{center}
\vspace{-0.1in}
\caption{Qualitative comparison of candidate-conditioned summarization from \name{} (Ours) compared to generic summarization as a rationale for the answer. More examples are in Appendix \ref{appendixC}.}
\vspace{-0.15in}
\label{fig:fig4_qualitative}
\end{figure*}

\textbf{Qualitative evaluation as rationale to answer.}
Lastly, we explore the additional benefits of \name{}, which offers rationales to support the prediction.
Specifically, we compare the summarization from \name{} with the generic summarization, which is also generated by LLMs but with no constraint of supporting specific answer candidates. 
To separately consider the quality as rationale with the accuracy of prediction, we only compare the samples correctly predicted by both \name{} and \textit{Generic summarization} used in Table \ref{table:ablation}; for example, it results in 84 remaining samples in the case of NQ$^{*}$. 
We first evaluate using GPT-4, which has been demonstrated to have a high correlation with humans \citep{liu2023gpteval}. 
We present the results in Figure \ref{fig:fig3b}.  
Here, one can observe that the summarization via \name{} is more preferred by GPT-4; for example, Generic summarization wins $30.3$\% while  \name{} wins $37.4$\% on average.
It is also worth noting that the average length of both summarizations is similar (Generic: 600 vs \name{}'s: 570 average characters on NQ), therefore the bias of GPT to prefer the longer response \citep{wang2023far} might limitedly affect the result. 
Next, we ask human evaluators \textit{which summarization is more informative and plausible to support the given question-answer pair} on 84 samples of NQ$^{*}$.
This result is presented in Figure \ref{fig:fig3c}. 
Here, we also observe a higher preference for \name{}'s summarization (Generic: $26.9$\% vs \name{}: $43.4$\%).
Overall, these results reveal the potential of \name{} toward a better ODQA system by providing a high-quality rationale for the answer.
Details on human evaluations are presented in Appendix \ref{appendixE}.

\section{Conclusion}

In this paper, we proposed \name{}, a simple yet effective framework to improve ODQA accuracy of LLMs.
Our key idea is to ensure the correctness of predicted answers by constructing the summaries of the retrieved passages for the potential answer candidates and evaluating their validity and ranking.
Our experiments demonstrate that \name{} significantly improves ODQA performance of various retrieval-augmented LLMs, and also has additional advantages for measuring the importance of passages and providing the rationale for prediction.
From these advantages, we believe our framework can contribute to various real-world applications and provide a better experience to users. 

\section*{Ethics Statement}
We strongly believe that \name{} can provide a strong positive impact in real-world applications related to QA, \textit{e.g.}, search engines or chatbots.
Since \name{} can provide the summarization that supports the corresponding prediction specifically, it can significantly improve the explainability \citep{mao2022explainable} and reliability \citep{whitehead2022reliable} of QA systems which are more important when they are constructed using black-box LLMs.
Moreover, considering the success of LLMs in various applications more than QA \citep{izacard2022few, nam2023semi}, we expect the advantages of this framework to better exploit the retrieved passages with LLMs will be beneficial to them. 

In contrast, there also exists some potential negative impacts when developing a system with the multiple usages of LLMs, as it could be costly \citep{chen2023frugalgpt} and generate sensitive \citep{santurkar2023whose} and malicious \citep{deshpande2023toxicity} text outputs. 
Since the summarization from \name{} is constructed based on the provided passages, one should consider their quality to prevent undesirable outputs.
On the other hand, incorporating the additional filtering could be a strong solution \citep{le2020adversarial, schick2021self}.
To reduce the cost, substituting specific steps of \name{}, \textit{e.g.}, measuring validity, with trainable small LMs could be an effective way, similar to \citet{yang2020generative, lewis2021paq, li2023symbolic}.

\section*{Reproducibility Statement}

We provide implementation details (\textit{e.g.}, design of prompts, used APIs, and retrieval methods) and experiment setups (\textit{e.g.}, datasets and metrics) in Section \ref{sec:4} and Appendix \ref{appendixB}.
In addition, we will release source codes near future. 

\section*{Acknowledgments}
This work was mainly supported by Institute of Information \& communications Technology Planning \& Evaluation (IITP) grant funded by the Korea government (MSIT) (No.2021-0-02068, Artificial Intelligence Innovation Hub; No.2019-0-00075, Artificial Intelligence Graduate School Program (KAIST)). This work was partly supported by KAIST-NAVER Hypercreative AI Center.

\bibliography{iclr2024_conference}

\begin{thebibliography}{74}
\providecommand{\natexlab}[1]{#1}
\providecommand{\url}[1]{\texttt{#1}}
\expandafter\ifx\csname urlstyle\endcsname\relax
  \providecommand{\doi}[1]{doi: #1}\else
  \providecommand{\doi}{doi: \begingroup \urlstyle{rm}\Url}\fi

\bibitem[Berant et~al.(2013)Berant, Chou, Frostig, and Liang]{berant2013semantic}
Jonathan Berant, Andrew Chou, Roy Frostig, and Percy Liang.
\newblock Semantic parsing on freebase from question-answer pairs.
\newblock In \emph{Conference on Empirical Methods in Natural Language Processing (EMNLP)}, 2013.

\bibitem[Borgeaud et~al.(2022)Borgeaud, Mensch, Hoffmann, Cai, Rutherford, Millican, Van Den~Driessche, Lespiau, Damoc, Clark, et~al.]{borgeaud2022improving}
Sebastian Borgeaud, Arthur Mensch, Jordan Hoffmann, Trevor Cai, Eliza Rutherford, Katie Millican, George~Bm Van Den~Driessche, Jean-Baptiste Lespiau, Bogdan Damoc, Aidan Clark, et~al.
\newblock Improving language models by retrieving from trillions of tokens.
\newblock In \emph{Proceedings of the International Conference on Machine Learning (ICML)}. PMLR, 2022.

\bibitem[Brown et~al.(2020)Brown, Mann, Ryder, Subbiah, Kaplan, Dhariwal, Neelakantan, Shyam, Sastry, Askell, et~al.]{brown2020language}
Tom Brown, Benjamin Mann, Nick Ryder, Melanie Subbiah, Jared~D Kaplan, Prafulla Dhariwal, Arvind Neelakantan, Pranav Shyam, Girish Sastry, Amanda Askell, et~al.
\newblock Language models are few-shot learners.
\newblock In \emph{Advances in Neural Information Processing Systems (NeurIPS)}, 2020.

\bibitem[Chen et~al.(2017)Chen, Fisch, Weston, and Bordes]{chen2017reading}
Danqi Chen, Adam Fisch, Jason Weston, and Antoine Bordes.
\newblock Reading wikipedia to answer open-domain questions.
\newblock In \emph{Annual Meeting of the Association for Computational Linguistics (ACL)}, 2017.

\bibitem[Chen et~al.(2023)Chen, Zaharia, and Zou]{chen2023frugalgpt}
Lingjiao Chen, Matei Zaharia, and James Zou.
\newblock Frugalgpt: How to use large language models while reducing cost and improving performance.
\newblock \emph{arXiv preprint arXiv:2305.05176}, 2023.

\bibitem[Chen et~al.(2021)Chen, Tworek, Jun, Yuan, Pinto, Kaplan, Edwards, Burda, Joseph, Brockman, et~al.]{chen2021evaluating}
Mark Chen, Jerry Tworek, Heewoo Jun, Qiming Yuan, Henrique Ponde de~Oliveira Pinto, Jared Kaplan, Harri Edwards, Yuri Burda, Nicholas Joseph, Greg Brockman, et~al.
\newblock Evaluating large language models trained on code.
\newblock \emph{arXiv preprint arXiv:2107.03374}, 2021.

\bibitem[Chowdhery et~al.(2022)Chowdhery, Narang, Devlin, Bosma, Mishra, Roberts, Barham, Chung, Sutton, Gehrmann, et~al.]{chowdhery2022palm}
Aakanksha Chowdhery, Sharan Narang, Jacob Devlin, Maarten Bosma, Gaurav Mishra, Adam Roberts, Paul Barham, Hyung~Won Chung, Charles Sutton, Sebastian Gehrmann, et~al.
\newblock Palm: Scaling language modeling with pathways.
\newblock \emph{arXiv preprint arXiv:2204.02311}, 2022.

\bibitem[Chowdhury(2010)]{chowdhury2010introduction}
Gobinda~G Chowdhury.
\newblock \emph{Introduction to modern information retrieval}.
\newblock Facet publishing, 2010.

\bibitem[Chuang et~al.(2023)Chuang, Fang, Li, Yih, and Glass]{chuang2023expand}
Yung-Sung Chuang, Wei Fang, Shang-Wen Li, Wen-tau Yih, and James Glass.
\newblock Expand, rerank, and retrieve: Query reranking for open-domain question answering.
\newblock In \emph{Findings of the Association for Computational Linguistics (ACL)}, 2023.

\bibitem[Creswell et~al.(2023)Creswell, Shanahan, and Higgins]{creswell2023selection}
Antonia Creswell, Murray Shanahan, and Irina Higgins.
\newblock Selection-inference: Exploiting large language models for interpretable logical reasoning.
\newblock In \emph{International Conference on Learning Representations (ICLR)}, 2023.

\bibitem[Deshpande et~al.(2023)Deshpande, Murahari, Rajpurohit, Kalyan, and Narasimhan]{deshpande2023toxicity}
Ameet Deshpande, Vishvak Murahari, Tanmay Rajpurohit, Ashwin Kalyan, and Karthik Narasimhan.
\newblock Toxicity in chatgpt: Analyzing persona-assigned language models.
\newblock \emph{arXiv preprint arXiv:2304.05335}, 2023.

\bibitem[Du et~al.(2022)Du, Huang, Dai, Tong, Lepikhin, Xu, Krikun, Zhou, Yu, Firat, et~al.]{du2022glam}
Nan Du, Yanping Huang, Andrew~M Dai, Simon Tong, Dmitry Lepikhin, Yuanzhong Xu, Maxim Krikun, Yanqi Zhou, Adams~Wei Yu, Orhan Firat, et~al.
\newblock Glam: Efficient scaling of language models with mixture-of-experts.
\newblock In \emph{Proceedings of the International Conference on Machine Learning (ICML)}, 2022.

\bibitem[Efron \& Tibshirani(1994)Efron and Tibshirani]{efron1994introduction}
Bradley Efron and Robert~J Tibshirani.
\newblock \emph{An introduction to the bootstrap}.
\newblock CRC press, 1994.

\bibitem[Gao et~al.(2023)Gao, Yen, Yu, and Chen]{gao2023enabling}
Tianyu Gao, Howard Yen, Jiatong Yu, and Danqi Chen.
\newblock Enabling large language models to generate text with citations.
\newblock In \emph{Conference on Empirical Methods in Natural Language Processing (EMNLP)}, 2023.

\bibitem[Giorgi et~al.(2023)Giorgi, Soldaini, Wang, Bader, Lo, Wang, and Cohan]{giorgi2023exploring}
John Giorgi, Luca Soldaini, Bo~Wang, Gary Bader, Kyle Lo, Lucy~Lu Wang, and Arman Cohan.
\newblock Open domain multi-document summarization: A comprehensive study of model brittleness under retrieval.
\newblock In \emph{Findings of the Conference on Empirical Methods in Natural Language Processing (EMNLP)}, 2023.

\bibitem[Guu et~al.(2020)Guu, Lee, Tung, Pasupat, and Chang]{guu2020retrieval}
Kelvin Guu, Kenton Lee, Zora Tung, Panupong Pasupat, and Mingwei Chang.
\newblock Retrieval augmented language model pre-training.
\newblock In \emph{International conference on machine learning}, pp.\  3929--3938. PMLR, 2020.

\bibitem[Hendrycks et~al.(2021)Hendrycks, Burns, Basart, Zou, Mazeika, Song, and Steinhardt]{hendrycks2020measuring}
Dan Hendrycks, Collin Burns, Steven Basart, Andy Zou, Mantas Mazeika, Dawn Song, and Jacob Steinhardt.
\newblock Measuring massive multitask language understanding.
\newblock In \emph{International Conference on Learning Representations (ICLR)}, 2021.

\bibitem[Ho et~al.(2020)Ho, Nguyen, Sugawara, and Aizawa]{ho2020constructing}
Xanh Ho, Anh-Khoa~Duong Nguyen, Saku Sugawara, and Akiko Aizawa.
\newblock Constructing a multi-hop qa dataset for comprehensive evaluation of reasoning steps.
\newblock \emph{arXiv preprint arXiv:2011.01060}, 2020.

\bibitem[Izacard et~al.(2022)Izacard, Caron, Hosseini, Riedel, Bojanowski, Joulin, and Grave]{izacard2021unsupervised}
Gautier Izacard, Mathilde Caron, Lucas Hosseini, Sebastian Riedel, Piotr Bojanowski, Armand Joulin, and Edouard Grave.
\newblock Unsupervised dense information retrieval with contrastive learning.
\newblock In \emph{Transactions on Machine Learning Research (TMLR)}, 2022.

\bibitem[Izacard et~al.(2023)Izacard, Lewis, Lomeli, Hosseini, Petroni, Schick, Dwivedi-Yu, Joulin, Riedel, and Grave]{izacard2022few}
Gautier Izacard, Patrick Lewis, Maria Lomeli, Lucas Hosseini, Fabio Petroni, Timo Schick, Jane Dwivedi-Yu, Armand Joulin, Sebastian Riedel, and Edouard Grave.
\newblock Few-shot learning with retrieval augmented language models.
\newblock In \emph{Journal of Machine Learning Research (JMLR)}, 2023.

\bibitem[Kamalloo et~al.(2023)Kamalloo, Dziri, Clarke, and Rafiei]{kamalloo2023evaluating}
Ehsan Kamalloo, Nouha Dziri, Charles~LA Clarke, and Davood Rafiei.
\newblock Evaluating open-domain question answering in the era of large language models.
\newblock In \emph{Annual Meeting of the Association for Computational Linguistics (ACL)}, 2023.

\bibitem[Karpukhin et~al.(2020)Karpukhin, Oguz, Min, Lewis, Wu, Edunov, Chen, and Yih]{karpukhin2020dense}
Vladimir Karpukhin, Barlas Oguz, Sewon Min, Patrick Lewis, Ledell Wu, Sergey Edunov, Danqi Chen, and Wen-tau Yih.
\newblock Dense passage retrieval for open-domain question answering.
\newblock In \emph{Conference on Empirical Methods in Natural Language Processing (EMNLP)}, 2020.

\bibitem[Kenton \& Toutanova(2019)Kenton and Toutanova]{kenton2019bert}
Jacob Devlin Ming-Wei~Chang Kenton and Lee~Kristina Toutanova.
\newblock Bert: Pre-training of deep bidirectional transformers for language understanding.
\newblock In \emph{Annual Conference of the North American Chapter of the Association for Computational Linguistics (NAACL)}, 2019.

\bibitem[Khattab et~al.(2021)Khattab, Potts, and Zaharia]{khattab2021baleen}
Omar Khattab, Christopher Potts, and Matei Zaharia.
\newblock Baleen: Robust multi-hop reasoning at scale via condensed retrieval.
\newblock In \emph{Advances in Neural Information Processing Systems (NeurIPS)}, 2021.

\bibitem[Kim et~al.(2021)Kim, Kim, Lee, Lee, Kwak, Hyeon, Park, Kim, Kim, Seo, et~al.]{kim2021changes}
Boseop Kim, HyoungSeok Kim, Sang-Woo Lee, Gichang Lee, Donghyun Kwak, Jeon~Dong Hyeon, Sunghyun Park, Sungju Kim, Seonhoon Kim, Dongpil Seo, et~al.
\newblock What changes can large-scale language models bring? intensive study on hyperclova: Billions-scale korean generative pretrained transformers.
\newblock In \emph{Conference on Empirical Methods in Natural Language Processing (EMNLP)}, 2021.

\bibitem[Kojima et~al.(2022)Kojima, Gu, Reid, Matsuo, and Iwasawa]{kojima2022large}
Takeshi Kojima, Shixiang~Shane Gu, Machel Reid, Yutaka Matsuo, and Yusuke Iwasawa.
\newblock Large language models are zero-shot reasoners.
\newblock In \emph{Advances in Neural Information Processing Systems (NeurIPS)}, 2022.

\bibitem[Kwiatkowski et~al.(2019)Kwiatkowski, Palomaki, Redfield, Collins, Parikh, Alberti, Epstein, Polosukhin, Devlin, Lee, et~al.]{kwiatkowski2019natural}
Tom Kwiatkowski, Jennimaria Palomaki, Olivia Redfield, Michael Collins, Ankur Parikh, Chris Alberti, Danielle Epstein, Illia Polosukhin, Jacob Devlin, Kenton Lee, et~al.
\newblock Natural questions: a benchmark for question answering research.
\newblock \emph{Transactions of the Association for Computational Linguistics}, 7:\penalty0 453--466, 2019.

\bibitem[Lazaridou et~al.(2022)Lazaridou, Gribovskaya, Stokowiec, and Grigorev]{lazaridou2022internet}
Angeliki Lazaridou, Elena Gribovskaya, Wojciech Stokowiec, and Nikolai Grigorev.
\newblock Internet-augmented language models through few-shot prompting for open-domain question answering.
\newblock \emph{arXiv preprint arXiv:2203.05115}, 2022.

\bibitem[Le~Bras et~al.(2020)Le~Bras, Swayamdipta, Bhagavatula, Zellers, Peters, Sabharwal, and Choi]{le2020adversarial}
Ronan Le~Bras, Swabha Swayamdipta, Chandra Bhagavatula, Rowan Zellers, Matthew Peters, Ashish Sabharwal, and Yejin Choi.
\newblock Adversarial filters of dataset biases.
\newblock In \emph{Proceedings of the International Conference on Machine Learning (ICML)}, 2020.

\bibitem[Lee et~al.(2019)Lee, Chang, and Toutanova]{lee2019latent}
Kenton Lee, Ming-Wei Chang, and Kristina Toutanova.
\newblock Latent retrieval for weakly supervised open domain question answering.
\newblock In \emph{Annual Meeting of the Association for Computational Linguistics (ACL)}, 2019.

\bibitem[Lewis et~al.(2021)Lewis, Wu, Liu, Minervini, K{\"u}ttler, Piktus, Stenetorp, and Riedel]{lewis2021paq}
Patrick Lewis, Yuxiang Wu, Linqing Liu, Pasquale Minervini, Heinrich K{\"u}ttler, Aleksandra Piktus, Pontus Stenetorp, and Sebastian Riedel.
\newblock Paq: 65 million probably-asked questions and what you can do with them.
\newblock \emph{Transactions of the Association for Computational Linguistics}, 9:\penalty0 1098--1115, 2021.

\bibitem[Li et~al.(2023)Li, Hessel, Yu, Ren, Chang, and Choi]{li2023symbolic}
Liunian~Harold Li, Jack Hessel, Youngjae Yu, Xiang Ren, Kai-Wei Chang, and Yejin Choi.
\newblock Symbolic chain-of-thought distillation: Small models can also" think" step-by-step.
\newblock In \emph{Annual Meeting of the Association for Computational Linguistics (ACL)}, 2023.

\bibitem[Liu et~al.(2023{\natexlab{a}})Liu, Jin, Wang, Cheng, Dou, and Wen]{liu2023reta}
Jiongnan Liu, Jiajie Jin, Zihan Wang, Jiehan Cheng, Zhicheng Dou, and Ji-Rong Wen.
\newblock Reta-llm: A retrieval-augmented large language model toolkit.
\newblock \emph{arXiv preprint arXiv:2306.05212}, 2023{\natexlab{a}}.

\bibitem[Liu et~al.(2023{\natexlab{b}})Liu, Lin, Hewitt, Paranjape, Bevilacqua, Petroni, and Liang]{liu2023lost}
Nelson~F Liu, Kevin Lin, John Hewitt, Ashwin Paranjape, Michele Bevilacqua, Fabio Petroni, and Percy Liang.
\newblock Lost in the middle: How language models use long contexts.
\newblock \emph{arXiv preprint arXiv:2307.03172}, 2023{\natexlab{b}}.

\bibitem[Liu et~al.(2023{\natexlab{c}})Liu, Iter, Xu, Wang, Xu, and Zhu]{liu2023gpteval}
Yang Liu, Dan Iter, Yichong Xu, Shuohang Wang, Ruochen Xu, and Chenguang Zhu.
\newblock Gpteval: Nlg evaluation using gpt-4 with better human alignment.
\newblock \emph{arXiv preprint arXiv:2303.16634}, 2023{\natexlab{c}}.

\bibitem[Mao et~al.(2022)Mao, Jiang, Wang, Liu, Xia, Lyu, and She]{mao2022explainable}
Jianguo Mao, Wenbin Jiang, Xiangdong Wang, Hong Liu, Yu~Xia, Yajuan Lyu, and Qiaoqiao She.
\newblock Explainable question answering based on semantic graph by global differentiable learning and dynamic adaptive reasoning.
\newblock In \emph{Conference on Empirical Methods in Natural Language Processing (EMNLP)}, 2022.

\bibitem[Mao et~al.(2021)Mao, He, Liu, Shen, Gao, Han, and Chen]{mao2021reader}
Yuning Mao, Pengcheng He, Xiaodong Liu, Yelong Shen, Jianfeng Gao, Jiawei Han, and Weizhu Chen.
\newblock Reader-guided passage reranking for open-domain question answering.
\newblock In \emph{Findings of the Association for Computational Linguistics (ACL)}, 2021.

\bibitem[Mialon et~al.(2023)Mialon, Dess{\`\i}, Lomeli, Nalmpantis, Pasunuru, Raileanu, Rozi{\`e}re, Schick, Dwivedi-Yu, Celikyilmaz, et~al.]{mialon2023augmented}
Gr{\'e}goire Mialon, Roberto Dess{\`\i}, Maria Lomeli, Christoforos Nalmpantis, Ram Pasunuru, Roberta Raileanu, Baptiste Rozi{\`e}re, Timo Schick, Jane Dwivedi-Yu, Asli Celikyilmaz, et~al.
\newblock Augmented language models: a survey.
\newblock In \emph{Transactions on Machine Learning Research (TMLR)}, 2023.

\bibitem[Min et~al.(2022)Min, Shi, Lewis, Chen, Yih, Hajishirzi, and Zettlemoyer]{min2022nonparametric}
Sewon Min, Weijia Shi, Mike Lewis, Xilun Chen, Wen-tau Yih, Hannaneh Hajishirzi, and Luke Zettlemoyer.
\newblock Nonparametric masked language modeling.
\newblock \emph{arXiv preprint arXiv:2212.01349}, 2022.

\bibitem[Nagel(2016)]{nagel2016common}
Sebastian Nagel.
\newblock Common crawl news dataset.
\newblock \emph{\url{https://data.commoncrawl.org/ crawl-data/CC-NEWS/index.html}}, 2016.

\bibitem[Nam et~al.(2023)Nam, Song, Park, Tack, Yun, Kim, and Shin]{nam2023semi}
Jaehyun Nam, Woomin Song, Seong~Hyeon Park, Jihoon Tack, Sukmin Yun, Jaehyung Kim, and Jinwoo Shin.
\newblock Semi-supervised tabular classification via in-context learning of large language models.
\newblock In \emph{Workshop on Efficient Systems for Foundation Models@ ICML2023}, 2023.

\bibitem[Nguyen et~al.(2016)Nguyen, Rosenberg, Song, Gao, Tiwary, Majumder, and Deng]{nguyen2016ms}
Tri Nguyen, Mir Rosenberg, Xia Song, Jianfeng Gao, Saurabh Tiwary, Rangan Majumder, and Li~Deng.
\newblock Ms marco: A human-generated machine reading comprehension dataset.
\newblock \emph{arXiv preprint arXiv:1611.09268}, 2016.

\bibitem[Ni et~al.(2022)Ni, Qu, Lu, Dai, {\'A}brego, Ma, Zhao, Luan, Hall, Chang, et~al.]{ni2022large}
Jianmo Ni, Chen Qu, Jing Lu, Zhuyun Dai, Gustavo~Hern{\'a}ndez {\'A}brego, Ji~Ma, Vincent~Y Zhao, Yi~Luan, Keith~B Hall, Ming-Wei Chang, et~al.
\newblock Large dual encoders are generalizable retrievers.
\newblock In \emph{Conference on Empirical Methods in Natural Language Processing (EMNLP)}, 2022.

\bibitem[OpenAI(2022)]{openai2022chatgpt}
OpenAI.
\newblock Introducing chatgpt.
\newblock \emph{\url{https://openai.com/blog/chatgpt}}, 2022.

\bibitem[OpenAI(2023)]{openai2023gpt4}
OpenAI.
\newblock Gpt-4 technical report.
\newblock \emph{arXiv preprint arXiv:2303.08774}, 2023.

\bibitem[Pillutla et~al.(2021)Pillutla, Swayamdipta, Zellers, Thickstun, Welleck, Choi, and Harchaoui]{pillutla2021mauve}
Krishna Pillutla, Swabha Swayamdipta, Rowan Zellers, John Thickstun, Sean Welleck, Yejin Choi, and Zaid Harchaoui.
\newblock Mauve: Measuring the gap between neural text and human text using divergence frontiers.
\newblock In \emph{Advances in Neural Information Processing Systems (NeurIPS)}, 2021.

\bibitem[Qin et~al.(2023)Qin, Jagerman, Hui, Zhuang, Wu, Shen, Liu, Liu, Metzler, Wang, et~al.]{qin2023large}
Zhen Qin, Rolf Jagerman, Kai Hui, Honglei Zhuang, Junru Wu, Jiaming Shen, Tianqi Liu, Jialu Liu, Donald Metzler, Xuanhui Wang, et~al.
\newblock Large language models are effective text rankers with pairwise ranking prompting.
\newblock \emph{arXiv preprint arXiv:2306.17563}, 2023.

\bibitem[Raffel et~al.(2020)Raffel, Shazeer, Roberts, Lee, Narang, Matena, Zhou, Li, and Liu]{raffel2020exploring}
Colin Raffel, Noam Shazeer, Adam Roberts, Katherine Lee, Sharan Narang, Michael Matena, Yanqi Zhou, Wei Li, and Peter~J Liu.
\newblock Exploring the limits of transfer learning with a unified text-to-text transformer.
\newblock \emph{Journal of Machine Learning Research (JMLR)}, 2020.

\bibitem[Rajpurkar et~al.(2016)Rajpurkar, Zhang, Lopyrev, and Liang]{rajpurkar2016squad}
Pranav Rajpurkar, Jian Zhang, Konstantin Lopyrev, and Percy Liang.
\newblock Squad: 100,000+ questions for machine comprehension of text.
\newblock In \emph{Conference on Empirical Methods in Natural Language Processing (EMNLP)}, 2016.

\bibitem[Reimers \& Gurevych(2019)Reimers and Gurevych]{reimers2019sentence}
Nils Reimers and Iryna Gurevych.
\newblock Sentence-bert: Sentence embeddings using siamese bert-networks.
\newblock In \emph{Conference on Empirical Methods in Natural Language Processing (EMNLP)}, 2019.

\bibitem[Robertson et~al.(2009)Robertson, Zaragoza, et~al.]{robertson2009probabilistic}
Stephen Robertson, Hugo Zaragoza, et~al.
\newblock The probabilistic relevance framework: Bm25 and beyond.
\newblock \emph{Foundations and Trends{\textregistered} in Information Retrieval}, 3\penalty0 (4):\penalty0 333--389, 2009.

\bibitem[Robinson et~al.(2023)Robinson, Rytting, and Wingate]{robinson2023leveraging}
Joshua Robinson, Christopher~Michael Rytting, and David Wingate.
\newblock Leveraging large language models for multiple choice question answering.
\newblock In \emph{International Conference on Learning Representations (ICLR)}, 2023.

\bibitem[Santurkar et~al.(2023)Santurkar, Durmus, Ladhak, Lee, Liang, and Hashimoto]{santurkar2023whose}
Shibani Santurkar, Esin Durmus, Faisal Ladhak, Cinoo Lee, Percy Liang, and Tatsunori Hashimoto.
\newblock Whose opinions do language models reflect?
\newblock In \emph{Proceedings of the International Conference on Machine Learning (ICML)}, 2023.

\bibitem[Schick et~al.(2021)Schick, Udupa, and Sch{\"u}tze]{schick2021self}
Timo Schick, Sahana Udupa, and Hinrich Sch{\"u}tze.
\newblock Self-diagnosis and self-debiasing: A proposal for reducing corpus-based bias in nlp.
\newblock \emph{Transactions of the Association for Computational Linguistics}, 9:\penalty0 1408--1424, 2021.

\bibitem[Shi et~al.(2023)Shi, Min, Yasunaga, Seo, James, Lewis, Zettlemoyer, and Yih]{shi2023replug}
Weijia Shi, Sewon Min, Michihiro Yasunaga, Minjoon Seo, Rich James, Mike Lewis, Luke Zettlemoyer, and Wen-tau Yih.
\newblock Replug: Retrieval-augmented black-box language models.
\newblock \emph{arXiv preprint arXiv:2301.12652}, 2023.

\bibitem[Si et~al.(2023)Si, Gan, Yang, Wang, Wang, Boyd-Graber, and Wang]{si2023prompting}
Chenglei Si, Zhe Gan, Zhengyuan Yang, Shuohang Wang, Jianfeng Wang, Jordan Boyd-Graber, and Lijuan Wang.
\newblock Prompting gpt-3 to be reliable.
\newblock In \emph{International Conference on Learning Representations (ICLR)}, 2023.

\bibitem[Stelmakh et~al.(2022)Stelmakh, Luan, Dhingra, and Chang]{stelmakh2022asqa}
Ivan Stelmakh, Yi~Luan, Bhuwan Dhingra, and Ming-Wei Chang.
\newblock Asqa: Factoid questions meet long-form answers.
\newblock In \emph{Conference on Empirical Methods in Natural Language Processing (EMNLP)}, 2022.

\bibitem[Su et~al.(2022{\natexlab{a}})Su, Li, Zhang, Shang, Jiang, Liu, and Fung]{su2022read}
Dan Su, Xiaoguang Li, Jindi Zhang, Lifeng Shang, Xin Jiang, Qun Liu, and Pascale Fung.
\newblock Read before generate! faithful long form question answering with machine reading.
\newblock In \emph{Findings of the Association for Computational Linguistics (ACL)}, 2022{\natexlab{a}}.

\bibitem[Su et~al.(2022{\natexlab{b}})Su, Lan, Wang, Yogatama, Kong, and Collier]{su2022contrastive}
Yixuan Su, Tian Lan, Yan Wang, Dani Yogatama, Lingpeng Kong, and Nigel Collier.
\newblock A contrastive framework for neural text generation.
\newblock In \emph{Advances in Neural Information Processing Systems (NeurIPS)}, 2022{\natexlab{b}}.

\bibitem[Sun et~al.(2023)Sun, Yan, Ma, Ren, Yin, and Ren]{sun2023chatgpt}
Weiwei Sun, Lingyong Yan, Xinyu Ma, Pengjie Ren, Dawei Yin, and Zhaochun Ren.
\newblock Is chatgpt good at search? investigating large language models as re-ranking agent.
\newblock \emph{arXiv preprint arXiv:2304.09542}, 2023.

\bibitem[Touvron et~al.(2023{\natexlab{a}})Touvron, Lavril, Izacard, Martinet, Lachaux, Lacroix, Rozi{\`e}re, Goyal, Hambro, Azhar, et~al.]{touvron2023llama1}
Hugo Touvron, Thibaut Lavril, Gautier Izacard, Xavier Martinet, Marie-Anne Lachaux, Timoth{\'e}e Lacroix, Baptiste Rozi{\`e}re, Naman Goyal, Eric Hambro, Faisal Azhar, et~al.
\newblock Llama: Open and efficient foundation language models.
\newblock \emph{arXiv preprint arXiv:2302.13971}, 2023{\natexlab{a}}.

\bibitem[Touvron et~al.(2023{\natexlab{b}})Touvron, Martin, Stone, Albert, Almahairi, Babaei, Bashlykov, Batra, Bhargava, Bhosale, et~al.]{touvron2023llama}
Hugo Touvron, Louis Martin, Kevin Stone, Peter Albert, Amjad Almahairi, Yasmine Babaei, Nikolay Bashlykov, Soumya Batra, Prajjwal Bhargava, Shruti Bhosale, et~al.
\newblock Llama 2: Open foundation and fine-tuned chat models.
\newblock \emph{arXiv preprint arXiv:2307.09288}, 2023{\natexlab{b}}.

\bibitem[Trivedi et~al.(2023)Trivedi, Balasubramanian, Khot, and Sabharwal]{trivedi2022interleaving}
Harsh Trivedi, Niranjan Balasubramanian, Tushar Khot, and Ashish Sabharwal.
\newblock Interleaving retrieval with chain-of-thought reasoning for knowledge-intensive multi-step questions.
\newblock In \emph{Annual Meeting of the Association for Computational Linguistics (ACL)}, 2023.

\bibitem[Voorhees et~al.(1999)]{voorhees1999trec}
Ellen~M Voorhees et~al.
\newblock The trec-8 question answering track report.
\newblock In \emph{Trec}, volume~99, pp.\  77--82, 1999.

\bibitem[Wang et~al.(2023)Wang, Ivison, Dasigi, Hessel, Khot, Chandu, Wadden, MacMillan, Smith, Beltagy, et~al.]{wang2023far}
Yizhong Wang, Hamish Ivison, Pradeep Dasigi, Jack Hessel, Tushar Khot, Khyathi~Raghavi Chandu, David Wadden, Kelsey MacMillan, Noah~A Smith, Iz~Beltagy, et~al.
\newblock How far can camels go? exploring the state of instruction tuning on open resources.
\newblock \emph{arXiv preprint arXiv:2306.04751}, 2023.

\bibitem[Wei et~al.(2022)Wei, Wang, Schuurmans, Bosma, Xia, Chi, Le, Zhou, et~al.]{wei2022chain}
Jason Wei, Xuezhi Wang, Dale Schuurmans, Maarten Bosma, Fei Xia, Ed~Chi, Quoc~V Le, Denny Zhou, et~al.
\newblock Chain-of-thought prompting elicits reasoning in large language models.
\newblock In \emph{Advances in Neural Information Processing Systems (NeurIPS)}, 2022.

\bibitem[Welleck et~al.(2020)Welleck, Kulikov, Roller, Dinan, Cho, and Weston]{welleck2020neural}
Sean Welleck, Ilia Kulikov, Stephen Roller, Emily Dinan, Kyunghyun Cho, and Jason Weston.
\newblock Neural text generation with unlikelihood training.
\newblock In \emph{International Conference on Learning Representations (ICLR)}, 2020.

\bibitem[Weng et~al.(2022)Weng, Zhu, He, Liu, and Zhao]{weng2022large}
Yixuan Weng, Minjun Zhu, Shizhu He, Kang Liu, and Jun Zhao.
\newblock Large language models are reasoners with self-verification.
\newblock \emph{arXiv preprint arXiv:2212.09561}, 2022.

\bibitem[Whitehead et~al.(2022)Whitehead, Petryk, Shakib, Gonzalez, Darrell, Rohrbach, and Rohrbach]{whitehead2022reliable}
Spencer Whitehead, Suzanne Petryk, Vedaad Shakib, Joseph Gonzalez, Trevor Darrell, Anna Rohrbach, and Marcus Rohrbach.
\newblock Reliable visual question answering: Abstain rather than answer incorrectly.
\newblock In \emph{International Conference on Computer Vision (ICCV)}, 2022.

\bibitem[Xuan-Quy et~al.(2023)Xuan-Quy, Ngoc-Bich, Xuan-Dung, Bac-Bien, and The-Duy]{xuan2023evaluation}
Dao Xuan-Quy, Le~Ngoc-Bich, Phan Xuan-Dung, Ngo Bac-Bien, and Vo~The-Duy.
\newblock Evaluation of chatgpt and microsoft bing ai chat performances on physics exams of vietnamese national high school graduation examination.
\newblock \emph{arXiv preprint arXiv:2306.04538}, 2023.

\bibitem[Yang et~al.(2020)Yang, Malaviya, Fernandez, Swayamdipta, Bras, Wang, Bhagavatula, Choi, and Downey]{yang2020generative}
Yiben Yang, Chaitanya Malaviya, Jared Fernandez, Swabha Swayamdipta, Ronan~Le Bras, Ji-Ping Wang, Chandra Bhagavatula, Yejin Choi, and Doug Downey.
\newblock Generative data augmentation for commonsense reasoning.
\newblock \emph{arXiv preprint arXiv:2004.11546}, 2020.

\bibitem[Yang et~al.(2018)Yang, Qi, Zhang, Bengio, Cohen, Salakhutdinov, and Manning]{yang2018hotpotqa}
Zhilin Yang, Peng Qi, Saizheng Zhang, Yoshua Bengio, William Cohen, Ruslan Salakhutdinov, and Christopher~D Manning.
\newblock Hotpotqa: A dataset for diverse, explainable multi-hop question answering.
\newblock In \emph{Conference on Empirical Methods in Natural Language Processing (EMNLP)}, 2018.

\bibitem[Zhao et~al.(2021)Zhao, Wallace, Feng, Klein, and Singh]{zhao2021calibrate}
Zihao Zhao, Eric Wallace, Shi Feng, Dan Klein, and Sameer Singh.
\newblock Calibrate before use: Improving few-shot performance of language models.
\newblock In \emph{Proceedings of the International Conference on Machine Learning (ICML)}, 2021.

\bibitem[Zhu et~al.(2015)Zhu, Kiros, Zemel, Salakhutdinov, Urtasun, Torralba, and Fidler]{zhu2015aligning}
Yukun Zhu, Ryan Kiros, Rich Zemel, Ruslan Salakhutdinov, Raquel Urtasun, Antonio Torralba, and Sanja Fidler.
\newblock Aligning books and movies: Towards story-like visual explanations by watching movies and reading books.
\newblock In \emph{European Conference on Computer Vision (ECCV)}, 2015.

\end{thebibliography}
\bibliographystyle{iclr2024_conference}

\appendix
\newpage
\section{Designed Prompts for Experiments}\label{appendixB}

In this section, we present the specific prompts used for the experiments in Section \ref{sec:4}.

\subsection{Answer candidates generation}
In Listing \ref{lst:answer_candidates_generation}, we present the prompt $p_{\tt can}$ which is used to generate $K$ answer candidates $\widetilde{\mathbf{y}}=[\widetilde{y}_{1}, \dots, \widetilde{y}_{K}]$ from the given question and $N$ retrieved passages (Eq. \ref{eq.2}).
Here, we present the case of $K=2$.
\begin{listing}[!ht]
\begin{minted}[fontsize=\footnotesize, frame=single, breaklines]{python}
f'''
Below are N passages related to the question at the end. After reading the passages, provide two correct candidates for the answer to the question at the end. Each answer should be in the form: (a) xx, (b) yy, and should not exceed 3 words for each candidate.

Passage #1 Title: {Passage #1 Title}
Passage #1 Text: {Passage #1 Text}

...

Passage #N Title: {Passage #N Title}
Passage #N Text: {Passage #N Text}

Question: {Question}

Answer:
'''
\end{minted}
\caption{Prompt for answer candidates generation.}
\label{lst:answer_candidates_generation}
\end{listing}

\newpage
\subsection{Conditional summarization}
In Listing \ref{lst:conditional_summarization}, we present the prompt $p_{\tt sum}$ which is used to generate conditional summarization $s_k$ of retrieved passages to validate each candidate $\widetilde{y}_{k}$ as an answer to the question (Eq. \ref{eq.3}). 
\begin{listing}[!ht]
\begin{minted}[fontsize=\footnotesize, frame=single, breaklines]{python}
f'''
Passage #1 Title: {Passage #1 Title}
Passage #1 Text: {Passage #1 Text}

...

Passage #N Title: {Passage #N Title}
Passage #N Text: {Passage #N Text}

Your job is to act as a professional writer. You will write a good-quality passage that can support the given prediction about the question only based on the information in the provided supporting passages.

Now, let's start. After you write, please write [DONE] to indicate you are done. Do not write a prefix (e.g., "Response:") while writing a passage.

Question: {Question}
Choices: {(a) Choice 1 (b) Choice 2}
Prediction: {(a) Choice 1 (or (b) Choice 2)}
Passage:
'''
\end{minted}
\caption{Prompt for conditional summarization.}
\label{lst:conditional_summarization}
\end{listing}

\subsection{Instance-wise validation}
In Listing \ref{lst:instance_validation}, we present the prompt $p_{\tt val}$ which is used to evaluate the validity of each summarization $s_k$ whether it is not a degenerated case as the provided passages are not enough to support $\widetilde{y}_{k}$, or it properly supports the given answer candidate $\widetilde{y}_{k}$, rather than the other candidate $\widetilde{y}_{i}, ~ i \neq k$ (Eq. \ref{eq.4}). 
\begin{listing}[!ht]
\begin{minted}[fontsize=\footnotesize, frame=single, breaklines]{python}
f'''
Question: {Question}

Prediction: {Prediction}

Passage: {Passage}

Does the passage correctly support the prediction? Choices: [True, False]. Answer:
'''
\end{minted}
\caption{Prompt for instance-wise validation.}
\label{lst:instance_validation}
\end{listing}

\newpage
\subsection{Pair-wise ranking}
In Listing \ref{lst:pair_wise_ranking}, we present the prompt $p_{\tt rank}$ which is used to evaluate how the given summarization $s_{k}$ is \textit{relatively informative} to answer the question $q$, among all summaries $S_{K}=\{s_{k}\}_{k=1}^{K}$ (Eq. \ref{eq.5}). 
\begin{listing}[!ht]
\begin{minted}[fontsize=\footnotesize, frame=single, breaklines]{python}
f'''
Question: Given the following passages, determine which one provides a more informative answer to the subsequent question.

Passage 1: {Passage 1}

Passage 2: {Passage 2}

Target Question: {Question}

Your Task:
Identify which passage (Passage 1 or Passage 2) is more relevant and informative to answer the question at hand. Choices: [Passage 1, Passage 2].

Answer:
'''
\end{minted}
\caption{Prompt for pair-wise ranking.}
\label{lst:pair_wise_ranking}
\end{listing}

\newpage
\subsection{Baseline prediction}
In Listing \ref{lst:baseline_prediction}, we present the prompt that is used to append the retrieved passages of the question to give it as inputs of LLMs. The result with this prompt is denoted by \textit{Base}, in Section \ref{sec:4}. 
The same prompt is used for \textit{no retrieval} by assuming $N=0$, \textit{i.e.}, the only question is given to LLMs with instruction.
\begin{listing}[!ht]
\begin{minted}[fontsize=\footnotesize, frame=single, breaklines]{python}
f'''
Passage #1 Title: {Passage #1 Title}
Passage #1 Text: {Passage #1 Text}

...

Passage #N Title: {Passage #N Title}
Passage #N Text: {Passage #N Text}

Task description: predict the answer to the following question. Do not exceed 3 words.

Question: {Question}

Answer:
'''
\end{minted}
\caption{Prompt for baseline prediction.}
\label{lst:baseline_prediction}
\end{listing}

\subsection{Prompts for general summarization}
In Listing \ref{lst:general_summarization}, we present the prompt that is used to construct \textit{generic} summarization used in Section \ref{sec:4.3}. One can observe that the conditioning part is removed, compared to $p_{\tt sum}$.
\begin{listing}[!ht]
\begin{minted}[fontsize=\footnotesize, frame=single, breaklines]{python}
f'''
Passage #1 Title: {Passage #1 Title}
Passage #1 Text: {Passage #1 Text}

...

Passage #N Title: {Passage #N Title}
Passage #N Text: {Passage #N Text}

Your job is to act as a professional writer. You will write a good-quality passage that can support the prediction about the question only based on the information in the provided supporting passages.

Now, let's start. After you write, please write [DONE] to indicate you are done. Do not write a prefix (e.g., "Response:") while writing a passage.

Question: {Question}
Passage:
'''
\end{minted}
\caption{Prompt for generic summarization.}
\label{lst:general_summarization}
\end{listing}

\subsection{Prompts for MCQ prompt}\label{A.7}

Recently, \citet{robinson2023leveraging} demonstrated that multiple-choice prompts generally elicit much more accurate responses than do cloze prompts, for LLMs with high multiple-choice symbol binding ability like OpenAI Codex \citep{chen2021evaluating}.
Motivated by this, we consider MCQ prompt in Listing \ref{lst:mcq_prompt} and use it in Table \ref{table:ablation}, to evaluate the effectiveness of selecting the answer from the construction and verification of the conditional summarizations rather than direct prompting, under the same answer candidates from Eq. \ref{eq.2}.
One can observe that the conditioning with multiple choices part is added, compared to baseline prompting in Listing 5.

\begin{listing}[!ht]
\begin{minted}[fontsize=\footnotesize, frame=single, breaklines]{python}
f'''
Passage #1 Title: {Passage #1 Title}
Passage #1 Text: {Passage #1 Text}

...

Passage #N Title: {Passage #N Title}
Passage #N Text: {Passage #N Text}

Task description: predict the answer to the following question. Do not exceed 3 words.

Question: {Question}
Choices: {(a) Choice 1 (b) Choice 2}
Answer:
'''
\end{minted}
\caption{Prompt for MCQ prompt.}
\label{lst:mcq_prompt}
\end{listing}

\subsection{Design principles for prompt}\label{A.8}

Before finalizing the prompts used in the experiments, we examined several prompt designs and chose the best-performing one. Here, we'd like to share two key observations from this process.
First, precise and detailed instructions are crucial. 
As each component of the proposed framework operates in a zero-shot manner, its output greatly relies on the provided instruction. 
For example, in answer candidate generation (Eq.~\ref{eq.2}), the current prompt, outlined in Listing \ref{lst:answer_candidates_generation}, consistently outperforms the initially considered simple prompt (\texttt{Task description: give two candidates for the answer to the following question (e.g., (a) xx, (b) yy)}).
Second, proper input arguments are essential. 
For instance, along with the target candidate, providing all candidates as additional input enhanced the quality of conditional summarization. 
This is because it further specifies which contexts of retrieval should be the focus. 
However, including this information, or even the retrieval passages, disrupted the verification step by interrupting the focus on the summarizations.

\section{Additional Quantitative Results}\label{appendixC}

\subsection{More results for \name{} under different configurations}
\label{appx:more_config}

In Table \ref{table:diff_retrieval_llm_f1}, we present F1 scores with different configurations of various LLMs and retrieval methods. 
Similar to the result in Table \ref{table:diff_retrieval_llm_em}, it is observed that \name{} consistently improves ODQA accuracy regardless
of types of LLMs and retrieval methods, with 3.2\% average F1 improvement on average.

\begin{table*}[t]
\vspace{-0.1in}
\centering
\caption{F1 with different configurations of LLMs and retrieval methods on four QA datasets. $N=10$ most relevant passages are commonly retrieved. For LLaMA2-chat, we conducted experiments on NQ$^{*}$ and WebQ$^{*}$ and the results are indicated by $^{*}$.
\vspace{0.1in}
}
\scalebox{0.8}{
{
\begin{tabular}{r|cccccc|cc|cc}
\toprule
& \multicolumn{6}{c|}{ChatGPT} & \multicolumn{2}{c|}{GPT-4} & \multicolumn{2}{c}{LLaMA2-chat} \\ \cmidrule(l){2-7} \cmidrule(l){8-11}
Datasets & BM25 & + \name{} & DPR & + \name{} & Contriever & + \name{} & BM25 & + \name{} & BM25 & + \name{} \\ \midrule

NQ & 38.8 & \textbf{42.3} & 47.4 & \textbf{50.8} & 47.6 & \textbf{50.4} & 40.9 & \textbf{42.4} & 36.4$^{*}$ & \textbf{42.9}$^{*}$ \\
WebQ & 32.6 & \textbf{36.6} & 37.5 & \textbf{40.4} & 37.7 & \textbf{40.9} & \textbf{36.4} & {32.1}  & 35.3$^{*}$ & \textbf{40.5}$^{*}$ \\ 
2Wiki & 32.8 & \textbf{38.1} & 22.8 & \textbf{25.2} & 32.9 & \textbf{37.1} & 39.2 & \textbf{43.2}  & 31.2 & \textbf{36.2} \\ 
HotpotQA & 40.3 & \textbf{43.4} & 34.6 & \textbf{35.5} & 42.9 & \textbf{43.5} & 44.4 & \textbf{50.4} & \textbf{39.6} & {38.4} \\ 
\midrule
Average & 36.1 & \textbf{40.1} & 35.6 & \textbf{38.0} & 40.3 & \textbf{43.0} & 40.2 & \textbf{42.0} & 35.6 & \textbf{39.5} \\ 
\bottomrule
\end{tabular}}
}
\label{table:diff_retrieval_llm_f1}
\end{table*}

\subsection{Limited achievable improvement with more candidates}

\begin{table*}[t]
\vspace{-0.1in}

\centering
\caption{
EM / F1 with different $K$ under ChatGPT. $N=10$ most relevant passages are commonly retrieved with BM25.
\vspace{0.1in}
}
\scalebox{0.95}{
{
\begin{tabular}{r|cccc|c}
\toprule
Datasets / Methods & NQ$^{*}$ & WebQ$^{*}$ & 2Wiki & HotpotQA & Average \\ \midrule
No retrieval & 26.4 / 37.9  & 20.4 / {36.7} & 21.4 / 24.8 & 23.2 / 34.8 & 22.9 / 33.6 \\ \midrule
Base & 29.4 / 41.7 & 19.4 / 32.2 & {27.4} / {32.8} & 30.8 / 40.3 & 27.0 / 36.9 \\
Oracle ($K=2$) & 43.0 / 53.9 & 29.0 / 43.9 
& 47.6 / 54.4 & 41.2 / 52.7  
& 40.2 / 51.2 \\
Oracle ($K=3$) & 45.2 / 56.0 & 29.8 / 47.2 
& 48.4 / 56.2 & 41.0 / 54.0  
& 41.1 / 53.4 \\
\bottomrule
\end{tabular}}
}
\label{table:supp_oracle_diff_k}
\end{table*}

As we denoted in Section \ref{sec:4.1}, we use a fixed value of $K=2$ for all the experiments. 
This is due to our initial observation that the room for improvement by increasing $K$ is not large compared to the additional costs.
To investigate this, we first assume the method, denoted \textit{Oracle}, which takes the maximum of EM and F1 among the multiple candidates, \textit{e.g.,}, if one candidate is true and the other is wrong, then \textit{Oracle} consider it as true.
As one can see in Table \ref{table:supp_oracle_diff_k}, increasing $K=3$ from $K=2$ limitedly improves the accuracy (\textit{e.g.}, 0.9\% in EM), compared to the remaining room for improvement by better selection with small $K$; for example, there is 9.0\% gap between \name{} and \textit{Oracle}, in terms of EM.
Therefore, in this work, we keep $K=2$ but we remark that \name{} can be extended with $K>2$. 
Also, as there is remaining room for improvement, we hope that future work could reduce such a gap. 

\subsection{Additional evaluation with LLMs}

In Section \ref{sec:4}, we considered EM/F1 scores as the common metrics for the considered ODQA datasets, following the previous works \citep{chowdhery2022palm, touvron2023llama1, izacard2022few, shi2023replug}, to make it easy to notice the significance of our results. 
Nevertheless, other factors like response coherence, relevance, and efficiency are important metrics to be considered. 

To evaluate these aspects, we have conducted additional evaluations with LLMs approaches. Specifically, we measured two additional metrics: (1) MAUVE \citep{pillutla2021mauve} and (2) LLM-acc \citep{kamalloo2023evaluating}. 
MAUVE is a recently proposed metric to compare the two distributions of the text generation model and human-written text using divergence frontiers. 
MAUVE (scale of 0 to 100, higher is better) is known for correlating highly with human judgments, and is frequently used to evaluate LMs’ responses \citep{su2022contrastive, gao2023enabling}. 
LLM-acc assesses the accuracy (\%) of LLMs’ responses to questions, using the prompting of LLMs instead of term overlap like EM/F1. We used the official code from the authors, only changing LLMs to ChatGPT.
We measured this metric on NQ$^{*}$, WebQ$^{*}$, 2Wiki, and HotpotQA datasets, and the results are presented in Table \ref{table:supp_llm_eval}.

\begin{table*}[t]
\vspace{-0.1in}
\centering
\caption{
MAUVE \citep{pillutla2021mauve} and LLM-evaluated accuracy \citep{kamalloo2023evaluating}. We use ChatGPT and $N=10$ most relevant passages are commonly retrieved with BM25.
}
\vspace{0.1in}
\scalebox{0.95}{
{
\begin{tabular}{r|cccc|c}
\toprule
MAUVE / LLM-acc & NQ$^{*}$ & WebQ$^{*}$ & 2Wiki & HotpotQA & Average \\ \midrule
Base & 81.3 / 53.2 & 61.3 / 48.8 & 35.1 / 36.2 & 62.4 / 51.6 & 60.0 / 47.5 \\ 
\name{} (Ours) & 95.9 / 56.2 & 75.7 / 51.4 & 52.2 / 48.2 & 89.6 / 52.4 & 78.3 / 52.1 \\
\bottomrule
\end{tabular}}
}
\label{table:supp_llm_eval}
\end{table*}

Here, it is observed that the proposed method also makes significant improvements compared to the baseline under these two additional evaluations with LLMs approaches. Along with the results in Section \ref{sec:4}, this result further validates that our framework enables LLMs to provide better answers to the given question. 

\subsection{Experiments on long-form question answering}

While we mainly conduct the experiments with QA datasets that have short answers in Section \ref{sec:4}, our approach has the potential to be applicable beyond short-answer datasets. To verify this, we have conducted additional experiments on long-form question answering tasks to validate our approach's applicability.
Specifically, we used ASQA dataset \citep{stelmakh2022asqa, gao2023enabling} which consists of factoid questions and the corresponding long-form answers; for example, the answers of ASQA dataset have an average length of 71.8 words, while the answers of NQ dataset have 2.6 words.
Following the setups in \citet{gao2023enabling}, we compared the base prompting method with retrieval and name{} (ours) on 948 test examples, using ChatGPT (\texttt{GPT-3.5-turbo-0301}) with 5 retrieved passages via GTR \citep{ni2022large} for the experiments. For the evaluation, we measure ROUGE-L and String Exact Match (STR-EM) for correctness, and MAUVE \citep{pillutla2021mauve} for fluency and coherence, following the previous works \citep{stelmakh2022asqa, gao2023enabling}. 

\begin{table*}[t]
\vspace{-0.1in}
\centering
\caption{
Evaluation on ASQA dataset \citep{stelmakh2022asqa}. We use ChatGPT and $N=5$ most relevant passages are commonly retrieved with GTR \citep{ni2022large}, following \citet{gao2023enabling}.
}
\vspace{0.1in}
\scalebox{0.95}{
{
\begin{tabular}{r|ccc}
\toprule
Methods / Metrics & ROUGE-L & STR-EM & MAUVE \\ \midrule
Base & 38.00 & 39.81 & 69.83 \\ 
\name{} (Ours) & 39.83 & 42.63 & 70.33 \\
\bottomrule
\end{tabular}}
}
\label{table:supp_long_qa}
\end{table*}

The results are presented in Table \ref{table:supp_long_qa}. 
One can observe that our proposed framework consistently improves the performance of retrieval-augmented LLMs for long-form QA tasks. However, we acknowledge that there is still room for improvement, particularly in finding better prompt designs, given that our current designs are based on performance on short-answer datasets. We hope future research will explore this direction, extending the benefits of our framework to broader QA scenarios with LLMs.

\subsection{Experimental with few-shot examples}\label{appendix}

\begin{table*}[t]
\vspace{-0.1in}
\centering
\caption{
Few-shot experimental results. We use ChatGPT and $N=10$ most relevant passages are commonly retrieved with BM25.
}
\vspace{0.1in}
\scalebox{0.95}{
{
\begin{tabular}{r|ccc}
\toprule
EM / F1 & 0-shot & 1-shot  & 5-shot \\ \midrule
NQ$^{*}$: Base & 29.4 / 41.7 & 30.1 / 39.3 & 31.9 / 42.0 \\ 
NQ$^{*}$: \name{} (Ours) & 35.6 / 44.9 & 36.3 / 46.8 & 37.2 / 47.7 \\ \midrule
WebQ$^{*}$: Base & 19.4 / 32.2 & 19.6 / 32.9 & 19.9 / 33.5 \\ 
WebQ$^{*}$: \name{} (Ours) & 23.2 / 36.5 & 24.2 / 39.4 & 24.3 / 38.5 \\ 
\bottomrule
\end{tabular}}
}
\label{table:supp_fewshot}
\end{table*}

Here, we conduct additional experiments on NQ$^{*}$ and WebQ$^{*}$, using 1-shot and 5-shot examples from training datasets during prediction. We compare the average EM/F1 of base prompting with retrieval and \name{}, across four different random seeds used for sample selection.
In Listing \ref{lst:few_shot_prompt}, we present the prompt that is used to generate $K$ answer candidates in the case where few-shot examples are given. Here, we present the case of $K=2$. Note that if few-shot examples are provided, only the prompt for generating answer candidates is modified.
Also, in Listing \ref{lst:few_shot_prompt_nq}, we present the prompt for the base prompting.
Table \ref{table:supp_fewshot} shows that adding few-shot examples improves QA accuracy for both the baseline and name{}. Specifically, we observed that name{}'s gain primarily results from generating more accurate answer candidates. These findings suggest that our proposed method could be effective in scenarios beyond the zero-shot setup considered. Therefore, we believe that our work could contribute to broader ODQA scenarios in the future.

\begin{listing}[!ht]
{
\begin{minted}[fontsize=\footnotesize, frame=single, breaklines]{python}
f'''
Below are N passages related to the question at the end. We also provide the answers for various questions. After reading the passages and question-answer pairs, provide two correct candidates for the answer to the question at the end. Each answer should be in the form: (a) xx, (b) yy, and should not exceed 3 words for each candidate.

Passage #1 Title: {Passage #1 Title}
Passage #1 Text: {Passage #1 Text}

...

Passage #N Title: {Passage #N Title}
Passage #N Text: {Passage #N Text}

Question: {Example question #1}
Answer: {Example answer #1}

...

Question: {Example question #shot}
Answer: {Example answer #shot}

Question: {Query question}
Provide two correct candidates for the answer:
'''
\end{minted}
}
\caption{{Prompt for answer candidates generation with few-shot examples.}}
{\label{lst:few_shot_prompt}}
\end{listing}

\begin{listing}[!ht]
{
\begin{minted}[fontsize=\footnotesize, frame=single, breaklines]{python}
f'''
Passage #1 Title: {Passage #1 Title}
Passage #1 Text: {Passage #1 Text}

...

Passage #N Title: {Passage #N Title}
Passage #N Text: {Passage #N Text}

Task description: predict the answer to the following question. Do not exceed 3 words.

Question: {Example question #1}
Answer: {Example answer #1}

...

Question: {Example question #K shot}
Answer: {Example answer #K shot}

Question: {Query question}
Answer:
'''
\end{minted}
}
\caption{Base prompt with few-shot examples.}
\label{lst:few_shot_prompt_nq}
\end{listing}

\section{Human Evaluation of Generated Summarization}\label{appendixE}

In this section, we provide details on the human preference evaluation of generated summarizations in Figure \ref{fig:fig3c}.
First, we generate summarizations with a generic method (Listing \ref{lst:general_summarization}) and with our proposed \name{} (Listing \ref{lst:conditional_summarization}).
To separately consider the quality as rationale with the accuracy of prediction, we only compare the samples correctly predicted by both \name{} and generic summarization; it results in 84 examples from the NQ$^{*}$.
Then, using the prompt in Listing \ref{lst:human_evaluation}, we conduct human evaluation.
Specifically, we hired seven NLP experts off-line for our human evaluation experiment. 
Unlike asking GPT-4 with Listing \ref{lst:gpt_based_evaluation}, we ask human evaluators to answer as a tie if it is hard to determine.

\begin{listing}[!ht]
\begin{minted}[fontsize=\footnotesize, frame=single, breaklines]{python}
f'''
Question: Given the following summaries for the target question, determine which one is more informative and plausible as rationale to support a given target question-answer pair.

Summary 1: {Summary 1}

Summary 2: {Summary 2}

Target Question: {Question}

Target Answer: {Answer}

Your Task:
Identify which summary (Summary 1 or Summary 2) is more informative and plausible as rationale to support a given answer at hand. Choices: [Summary 1, Summary 2].

Answer:
'''
\end{minted}
\caption{Prompt for GPT-based evaluation.}
\label{lst:gpt_based_evaluation}
\end{listing}
\begin{listing}[!ht]
\begin{minted}[fontsize=\footnotesize, frame=single, breaklines]{python}
f'''
Given the following summaries for the target question, determine which one is more informative and plausible as rationale to support a given target question-answer pair.

Target Question: {Question}

Target Answer: {Answer}

Summary 1: {Summary 1}

Summary 2: {Summary 2}

Choices: [Summary 1, Tie, Summary 2]

Your choice: 
'''
\end{minted}
\caption{Template for human evaluation.}
\label{lst:human_evaluation}
\end{listing}

\newpage
\section{Additional Qualitative Results}\label{appendixD}

In this section, we present more qualitative results with \name{}. 
All the examples are from NQ$^{*}$, and ChatGPT with BM25 ($N=10$) is commonly used. 

\subsection{More examples of qualitative comparison between \name{}'s summarization and generic summarization}

\begin{figure*}[t]
\vspace{-0.1in}
\begin{center}
    \includegraphics[width=0.8\textwidth]{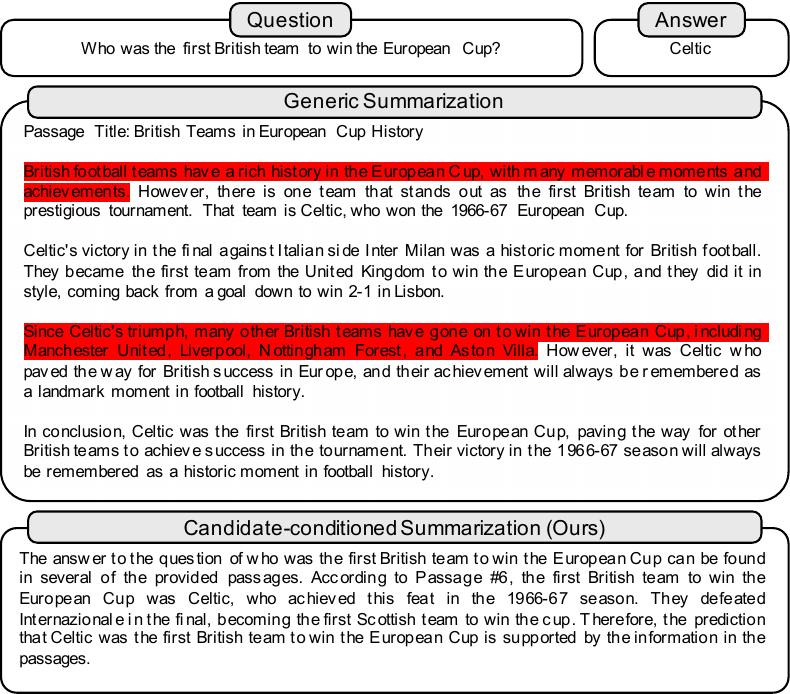}
\end{center}
\vspace{0.2in}
\begin{center}
    \includegraphics[width=0.8\textwidth]{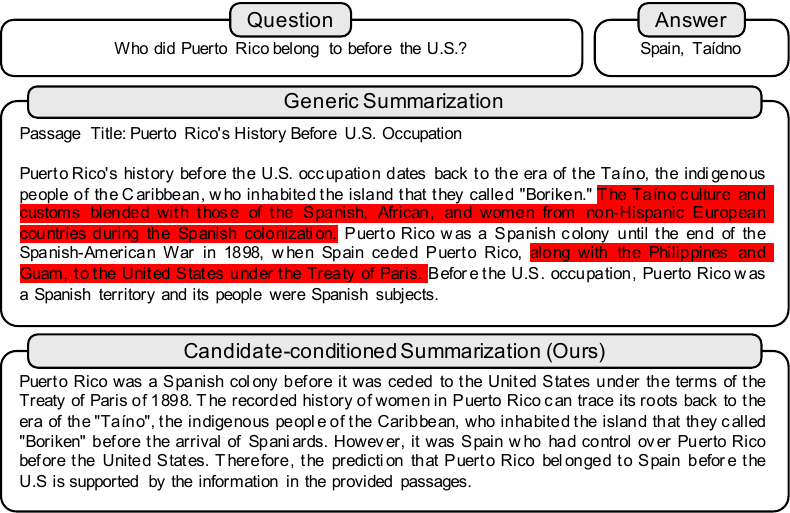}
\end{center}
\vspace{-0.1in}
\caption{Qualitative comparison of candidate-conditioned summarization from \name{} (Ours) compared to generic summarization as a rationale for the answer.}
\vspace{-0.1in}
\label{fig:supp_qualitative1}
\end{figure*}

\begin{figure*}[t]
\vspace{-0.1in}
\begin{center}
    \includegraphics[width=0.8\textwidth]{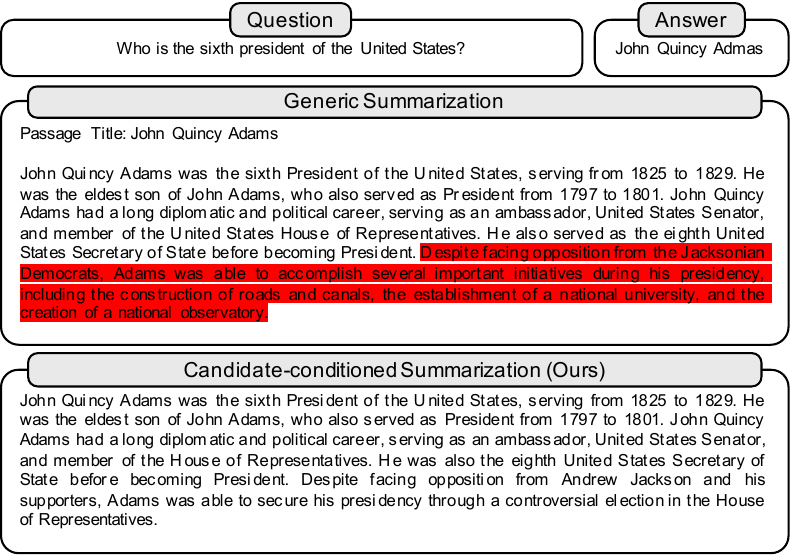}
\end{center}
\vspace{0.2in}
\begin{center}
    \includegraphics[width=0.8\textwidth]{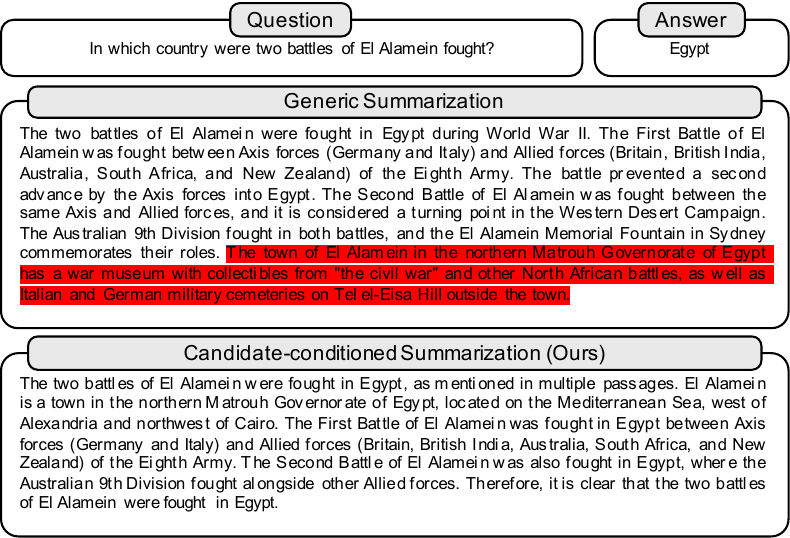}
\end{center}
\vspace{-0.1in}
\caption{Qualitative comparison of candidate-conditioned summarization from \name{} (Ours) compared to generic summarization as a rationale for the answer.}
\vspace{-0.1in}
\label{fig:supp_qualitative2}
\end{figure*}

\begin{figure*}[t]
\vspace{-0.1in}
\begin{center}
    \includegraphics[width=0.8\textwidth]{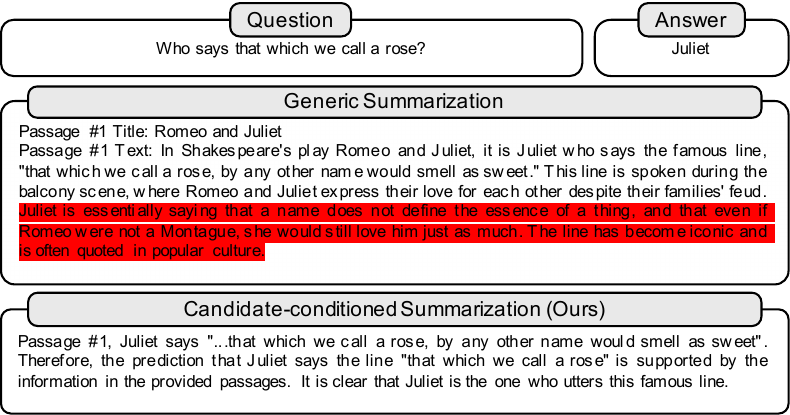}
\end{center}
\vspace{0.2in}
\begin{center}
    \includegraphics[width=0.8\textwidth]{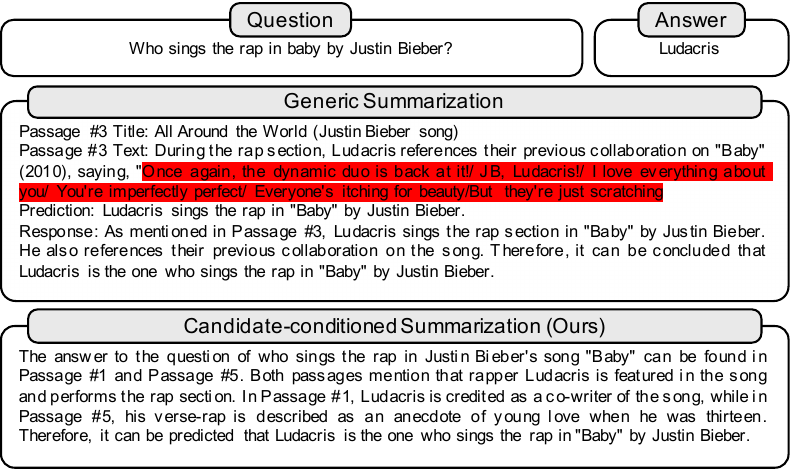}
\end{center}
\vspace{-0.1in}
\caption{Qualitative comparison of candidate-conditioned summarization from \name{} (Ours) compared to generic summarization as a rationale for the answer.}
\vspace{-0.1in}
\label{fig:supp_qualitative3}
\end{figure*}

In Figures \ref{fig:supp_qualitative1}, \ref{fig:supp_qualitative2}, and \ref{fig:supp_qualitative3}, we present more examples for qualitative comparison between the candidate-conditioned summarization by \name{} and generic summarization.
Innecessary and tedious sentences irrelevant to the answer are highlighted with \textbf{\textcolor{red}{red}}.

\subsection{Qualitative examples of verification with instance-wise validity}

\begin{figure*}[t]
\vspace{-0.1in}
\begin{center}
    \includegraphics[width=0.8\textwidth]{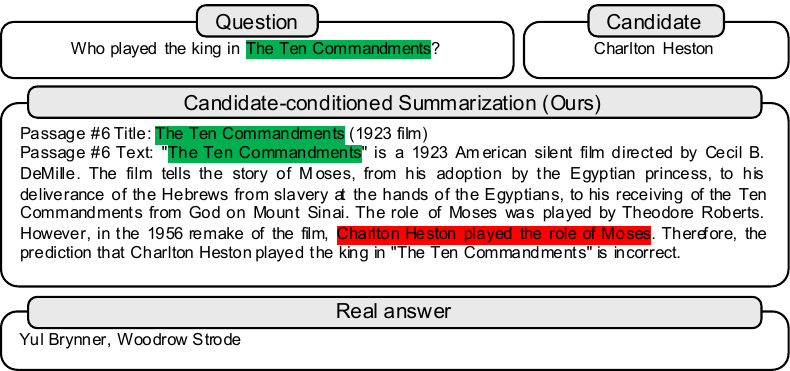}
\end{center}
\vspace{0.2in}
\begin{center}
    \includegraphics[width=0.8\textwidth]{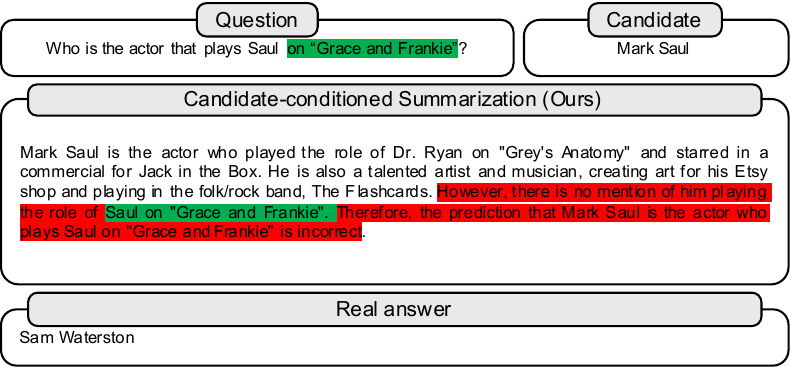}
\end{center}
\vspace{0.2in}
\begin{center}
    \includegraphics[width=0.8\textwidth]{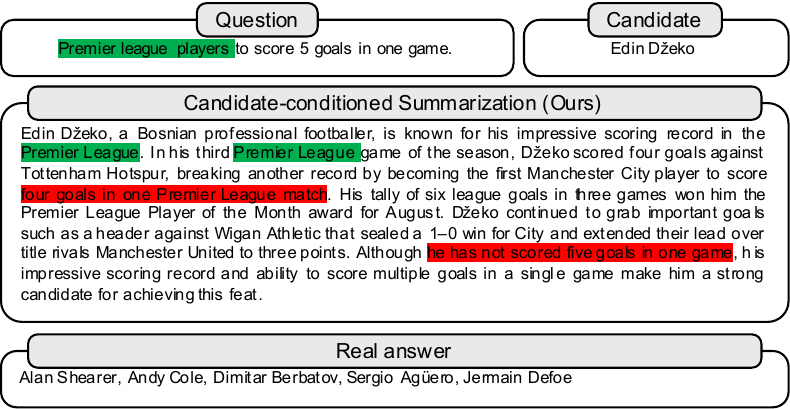}
\end{center}
\vspace{-0.1in}
\caption{Example summarizations that are evaluated as invalid by LLMs (Eq. \ref{eq.4}).}
\vspace{-0.1in}
\label{fig:supp_valid}
\end{figure*}

To qualitatively show which samples are considered as \textit{invalid} by LLMs, we present the examples that exhibit $v(s_{k}) = 0$ as $\mathcal{M}\left(p_{\tt val}(q, y_{k}, s_{k})\right) = \text{False}$  
in Figure \ref{fig:supp_valid}. 
Here, we highlight the sentences with \textbf{\textcolor{green}{green}} if they include the relevant context with the given candidate. 
In addition, we highlight the sentences with \textbf{\textcolor{red}{red}} if they induce a different candidate as an answer or do not support the candidate.  
For example, in the second example with a question (\texttt{Who is the actor that plays Saul on ``Grace and Frankie''?}), one can observe that the generated summarization concludes that the given candidate (\texttt{Mark Saul}) is incorrect; consequently, LLMs evaluates its validity as supporting summarization for the given candidate as false. 

\subsection{Qualitative examples of verification with pair-wise ranking}

\begin{figure*}[t]
\vspace{-0.1in}
\begin{center}
    \includegraphics[width=0.8\textwidth]{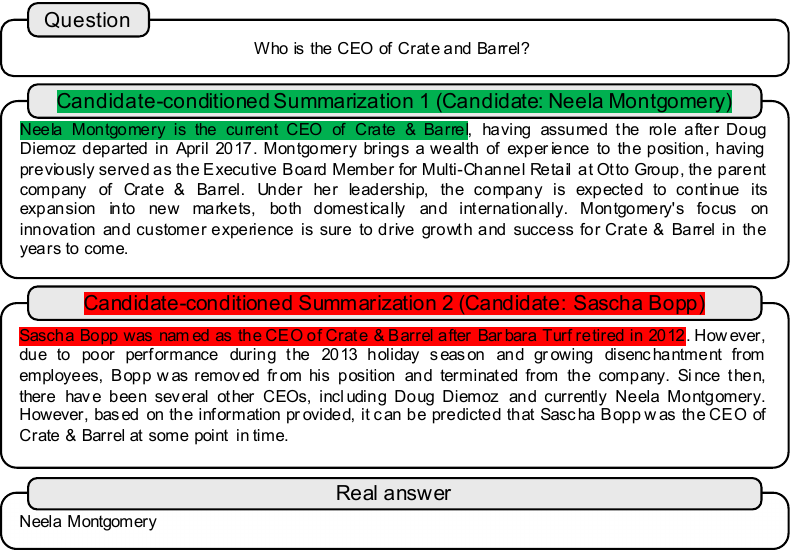}
\end{center}
\vspace{0.2in}
\begin{center}
    \includegraphics[width=0.8\textwidth]{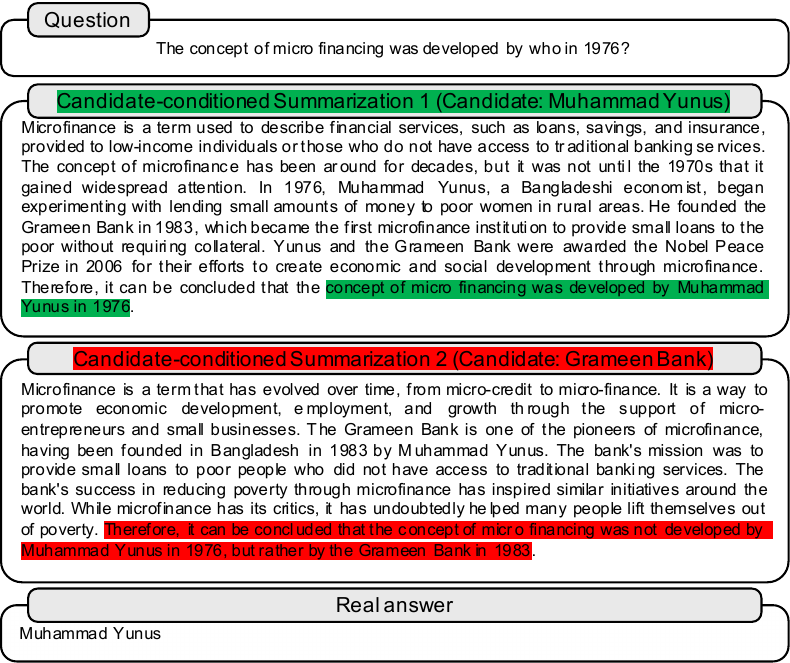}
\end{center}
\vspace{-0.1in}
\caption{Example summarizations with pair-wise rank evaluation (Eq. \ref{eq.5}).}
\vspace{-0.1in}
\label{fig:supp_rank}
\end{figure*}

In Figure \ref{fig:supp_rank}, we present examples of verification by pair-wise ranking. Here, we highlight with \textbf{\textcolor{green}{green}} for the summarization that gets a higher ranking.
In contrast,  we highlight with \textbf{\textcolor{red}{red}} for the summarization that gets a lower ranking.
We also highlight the relevant texts with the same colors, respectively.    

\section{Discussion on Cost and Quality Gain}

While \name{} significantly improves QA system of LLMs, one can be concerned about its cost as it requires multiple inferences of LLMs. 
However, we note that the improvement of \name{} is not just a simple consequence of more cost. 
Compared to other cost-increasing methods for accuracy improvement, \name{} significantly outperforms them, \textit{i.e.}, \name{} is an even more efficient way to increase performance. 
For instance, increasing the number of retrieved passages is one of the most straightforward methods for this goal. 
But, in this case, \name{} with 10 passages outperforms the base prompting with 50 passages, even with a lower total cost, as presented in Table \ref{tab:supp_cost}. 
In addition, we note that other baseline approaches such as chain-of-thought or self-verification (considered in Table \ref{table:main_odqa}) also require more cost than base prompting, but they fail to successfully improve the performance.

\begin{table*}[t]
    \centering
    \caption{Accuracy and cost (\$) for each method. For the method in the last row, ChatGPT is used for Eq \ref{eq.2} and \ref{eq.3}, and LLaMA is used for Eq \ref{eq.4} and \ref{eq.5}, respectively.}
    \vspace{0.1in}
    \scalebox{0.95}{
    \begin{tabular}{c|cc|c} 
    \toprule
         Exact Match (EM) & {NQ*} & {WebQ*} & {Average Cost for API} \\\midrule
         \text{Base (10 passages, ChatGPT)} & 29.4 & 19.4 & 1.57\$ \\
\text{Base (50 passages, ChatGPT)} & 33.8 & 21.8 & \textbf{7.67}\$ \\ 
\text{\name{} (10 passages, ChatGPT)} & \textbf{35.6} & 23.2 & 6.05\$ \\ \midrule
\text{\name{} (10 passages, ChatGPT + LLaMA)} & 35.0 & \textbf{24.8} & 5.03\$ \\ \bottomrule
\end{tabular}}\label{tab:supp_cost}
\end{table*}

On the other hand, one can reduce the overall cost by using cheaper LLMs for specific components, thanks to the modularity of \name{}. 
Remarkably, \name{} is compatible with the recent state-of-the-art open LLMs (see Tables \ref{table:diff_retrieval_llm_em} and \ref{table:diff_retrieval_llm_f1}) and hence this advantage is more noticeable. 
To give an intuition, we conduct the new experiments by using ChatGPT for the answer candidate generation and summarization, and LLaMA for the succeeding verification steps. 
As shown in the 4th row of Table \ref{tab:supp_cost}, this hybrid approach of different LLMs with \name{} successfully reduces the cost while keeping the effectiveness for improving the accuracy; for WebQ*, this approach even outperforms the expensive one. 
This result is from the effectiveness of LLaMA in WebQ* and indicates the potential of such a hybrid method. 

Lastly, we further remark that most of \name{}'s cost is currently from re-reading retrieved passages for conditional summarizations (\textit{e.g.}, 38\% of the total cost for \name{} with 10 passages). This is due to current APIs not providing recycling options for previous inputs. If recycling becomes available, \name{}'s cost could be significantly reduced. 

\section{Limitation and Future Work}
In this work, we primarily focused on zero-shot setup for the experiments, which is a commonly encountered scenario in the real world, \textit{e.g.}, search engine.
But, similar to the previous works \citep{chowdhery2022palm, touvron2023llama1}, incorporating data-specific few-shot examples is also an interesting future direction to further improve QA accuracy of LLMs with \name{}.
Another interesting direction is extending the applied task beyond QA, such as language modeling \citep{guu2020retrieval} or language understanding tasks \citep{hendrycks2020measuring}.

\section{Additional Related Work}\label{app:related_work}
\textbf{Summarization in open-domain.} A summarization of retrieved passages has been considered in open-domain context; for example, there are recent works that propose to learn a module to selectively use the retrieved information in sentence- 
\citep{khattab2021baleen, su2022read} or passage-level \citep{mao2021reader, chuang2023expand}.
In addition, \citet{su2022read, giorgi2023exploring} form a new task that combines both passage retrieval and summarization for a given query, and \citet{gao2023enabling} considers summarization of information for prompting.  
However, these works require a large annotated dataset to extract the information specified to answer the question or construct the generic summarization which focuses on preserving the retrieved information within reduced texts.  

\section{Experimental Results with Confidence Interval}

In this section, we present confidence intervals for our main tables (Tables \ref{table:main_odqa} and \ref{table:diff_retrieval_llm_em}). 
To achieve this, we apply bootstrapping~\citep{efron1994introduction}, a popular technique for statistical inference that involves random sampling with replacement. 
We report 95\% confidence intervals obtained through 1,000 iterations of bootstrapping. 
The confidence intervals for the EM and F1 metrics of each main table can be found in Tables \ref{table:supp_odqa_em_ci}, \ref{table:supp_odqa_f1_ci}, \ref{table:supp_diff_llm_em_ci}, and \ref{table:supp_diff_llm_f1_ci}.

The reliability of the results is reasonably robust, with the 95\% confidence interval having only about a 10\% variance from the reported value. 
Specifically, in the EM metric of the NQ dataset, our SuRe has the lowest confidence interval value at 32.0, compared to the maximum value of 29.1 for the no retrieval baseline and 30.0 for the best competitor. 
This demonstrates that the advantage of SuRe over prior works is statistically significant.

\begin{table*}[t]
\centering
{
\caption{EM with different QA methods with ChatGPT on four QA datasets. The 95\% confidence intervals are calculated via bootstrapping by 1000 iterations, and presented below the corresponding values. $N=10$ most relevant passages are retrieved using 
BM25, except \textit{no retrieval}. The best and second best scores are highlighted in \textbf{bold} and \underline{underline}, respectively.
\vspace{0.1in}
}
\scalebox{0.95}{
{
\begin{tabular}{r|cccc}
\toprule
Methods / Datasets & NQ & WebQ & 2Wiki & HotpotQA \\
\midrule
No retrieval & 27.6  & \underline{25.0}  & 21.4  & 22.2 \\ 
& {\small [26.2, 29.1]} & {\small[23.2, 27.0] } & {\small[17.6, 25.2] } & {\small[18.8, 25.8] }\\ \midrule
Base & \underline{28.4} & 19.6  & \underline{27.4} & 30.8  \\
& {\small[27.0, 30.0] }& {\small[17.9, 21.3] } & {\small[23.6, 31.0]}  & {\small [26.6, 35.2]} \\ 
Rerank & 24.8 & 18.8 & 23.0 & 27.8 \\
& {\small[23.4, 26.2] }& {\small[17.2, 20.6] } &{\small [19.6, 26.8]  }& {\small[23.8, 32.0]} \\ 
RePlug & 26.0 & 18.8 & 23.6 & 28.0 \\
& {\small[24.6, 27.4] }& {\small[17.1, 20.6]}  &{\small [20.0, 27.2]}  & {\small[24.2, 32.0]}  \\ 
Selection-inference & 24.3 & 17.3 & 22.6 & 30.8  \\
& {\small[22.9, 25.7] }& {\small[15.7, 18.8]  }& {\small[19.0, 26.0]}  & {\small[26.8, 34.8] } \\ 
Chain-of-thoughts 
& 22.3 & 15.2 & 19.6 & 25.6  \\ 
& {\small[20.8, 23.6] }& {\small[13.7, 16.6] } &{\small [16.0, 23.2] } & {\small[21.6, 29.4]} \\ 
Self-verification & 25.2 & 16.1 & 23.2 & \underline{31.6} \\ 
& {\small[23.7, 26.6] }& {\small[14.6, 17.7]}  &{\small [19.6, 27.6] } & {\small[27.6, 35.8]} \\ 
\midrule
\name{} (Ours) & \textbf{33.5} & \textbf{25.1} & \textbf{32.8} & \textbf{33.2}  \\
& {\small[32.0, 35.0]} &{\small [23.1, 27.0]}  &{\small [28.6, 36.8]}  & {\small[29.0, 37.6] } \\ 
\bottomrule
\end{tabular}}
}
\label{table:supp_odqa_em_ci}
}
\end{table*}
\begin{table*}[t]
\centering
{
\caption{F1 with different QA methods with ChatGPT on four QA datasets. The 95\% confidence intervals are calculated via bootstrapping by 1000 iterations, and presented below the corresponding values. $N=10$ most relevant passages are retrieved using 
BM25, except \textit{no retrieval}. The best and second best scores are highlighted in \textbf{bold} and \underline{underline}, respectively.
\vspace{0.1in}
}
\scalebox{0.95}{
{
\begin{tabular}{r|cccc}
\toprule
Methods / Datasets & NQ & WebQ & 2Wiki & HotpotQA \\
\midrule
No retrieval & \underline{39.0} & \textbf{38.8} & 24.8 & 31.9  \\ 
& [37.7, 40.5] & {\small[36.9, 40.7] } &{\small [21.1, 28.5]}  & {\small[28.2, 35.4]  }\\ \midrule
Base & 38.8 & 32.5 & \underline{32.8} & 40.3  \\
& [37.3, 40.4] & {\small[30.7, 34.3] } &{\small [28.9, 36.3]}  & {\small[36.4, 44.4]  }  \\ 
Rerank & 33.9 & 30.6 & 28.4 & 37.4 \\
& [32.5, 35.3] & {\small[28.8, 32.4] } &{\small [25.1, 32.1]}  & {\small[33.4, 41.4]  }  \\ 
RePlug & 35.3 & 31.5 & 28.5 & 37.9 \\
& [33.9, 36.8] & {\small[29.8, 33.3] } &{\small [24.7, 32.1]}  & {\small[34.2, 41.9]  }  \\ 
Selection-inference & 32.8 & 28.6 & 29.5 & 39.6 \\
& [31.5, 34.3] & {\small[27.0, 30.3] } &{\small [26.2, 33.2]}  & {\small[35.8, 43.7]  } \\ 
Chain-of-thoughts 
& 31.4 & 27.8 & 22.5 & 31.8  \\ 
& [30.1, 32.8] & {\small[26.1, 29.4] } &{\small [18.8, 26.2]}  & {\small[27.8, 35.7]  } \\ 
Self-verification & 35.4 & 28.5 & 30.5 & \underline{41.8}  \\ 
& [33.9, 36.9] & {\small[26.7, 30.2] } &{\small [27.1, 34.9]}  & {\small[37.8, 45.6]  }\\ 
\midrule
\name{} (Ours) & \textbf{42.3} &  \underline{36.6} & \textbf{38.1} & \textbf{43.4}  \\
& [40.8, 43.7] & {\small[34.8, 38.5] } &{\small [34.0, 42.0]}  & {\small[39.4, 47.6]  } \\ 
\bottomrule
\end{tabular}}
}
\label{table:supp_odqa_f1_ci}
}
\end{table*}
\begin{table*}[t]
\centering
{
\caption{EM with different configurations of LLMs and retrieval methods on four QA datasets.
The 95\% confidence intervals are calculated via bootstrapping by 1000 iterations, and presented below the corresponding values.
$N=10$ most relevant passages are commonly retrieved. For LLaMA2-chat, we conducted experiments on NQ$^{*}$ and WebQ$^{*}$ and the results are indicated by $^{*}$.
\vspace{0.1in}
}
\scalebox{0.7}{
{
\begin{tabular}{r|cccccc|cc|cc}
\toprule
& \multicolumn{6}{c|}{ChatGPT} & \multicolumn{2}{c|}{GPT-4} & \multicolumn{2}{c}{LLaMA2-chat} \\ \cmidrule(l){2-7} \cmidrule(l){8-11}
Datasets & BM25 & + \name{} & DPR & + \name{} & Contriever & + \name{} & BM25 & + \name{} & BM25 & + \name{} \\ \midrule

NQ & 28.4 & \textbf{33.5}  & 36.1 & \textbf{41.0} & 35.8 & \textbf{40.4} & 30.2 & \textbf{32.4}  & 18.6$^{*}$ & \textbf{30.4}$^{*}$ \\
& {\small [27.0, 30.0]} & {\small [32.0, 35.0]}& {\small [34.5, 37.7]}& {\small [39.4, 42.6]}& {\small [34.2, 37.4]}& {\small [38.8, 42.0]}& {\small [28.8, 31.7]}& {\small [30.9, 33.9]}& {\small [15.0, 22.0]}& {\small [26.4, 34.4]}\\ 
WebQ & 19.6 & \textbf{25.1} & 23.2 & \textbf{27.3} & 22.5 & \textbf{28.7} & 21.5 & \textbf{21.7}  & 16.0$^{*}$ & \textbf{24.0}$^{*}$ \\ 
& {\small [17.8, 21.4]} & {\small [23.1, 27.0]}& {\small [21.4, 25.1]}& {\small [25.3, 29.2]}& {\small [20.5, 24.3]}& {\small [26.7, 30.7]}& {\small [19.6, 23.2]}& {\small [19.8, 23.4]}& {\small [13.0, 19.2]}& {\small [20.4, 27.8]}\\
2Wiki & 27.4 & \textbf{32.8} & 19.2 & \textbf{21.4} & 27.2 & \textbf{32.6} & 34.8 & \textbf{38.2}  & 20.2 & \textbf{27.8} \\ 
& {\small [23.6, 31.0]} & {\small [28.6, 36.8]}& {\small [15.8, 22.8]}& {\small [17.8, 24.8]}& {\small [23.2, 31.0]}& {\small [28.4, 36.8]}& {\small [30.6, 38.8]}& {\small [34.0, 42.2]}& {\small [16.6, 23.6]}& {\small [23.6, 32.0]}\\
HotpotQA & 30.8 & \textbf{33.2} & 25.6 & \textbf{27.4} & 32.2 & \textbf{33.6} & 34.8 & \textbf{40.6}  & 24.0 & \textbf{28.0} \\ 
& {\small [26.6, 35.2]} & {\small [29.0, 37.6]}& {\small [21.8, 29.4]}& {\small [23.8, 31.4]}& {\small [28.2, 36.6]}& {\small [29.2, 37.6]}& {\small [31.0, 38.8]}& {\small [36.2, 44.6]}& {\small [20.2, 27.6]}& {\small [24.0, 32.0]}\\
\bottomrule
\end{tabular}}
}
\label{table:supp_diff_llm_em_ci}
}
\end{table*}

\begin{table*}[t]
\centering
{
\caption{F1 with different configurations of LLMs and retrieval methods on four QA datasets.
The 95\% confidence intervals are calculated via bootstrapping by 1000 iterations, and presented below the corresponding values.
$N=10$ most relevant passages are commonly retrieved. For LLaMA2-chat, we conducted experiments on NQ$^{*}$ and WebQ$^{*}$ and the results are indicated by $^{*}$.
\vspace{0.1in}
}
\scalebox{0.7}{
{
\begin{tabular}{r|cccccc|cc|cc}
\toprule
& \multicolumn{6}{c|}{ChatGPT} & \multicolumn{2}{c|}{GPT-4} & \multicolumn{2}{c}{LLaMA2-chat} \\ \cmidrule(l){2-7} \cmidrule(l){8-11}
Datasets & BM25 & + \name{} & DPR & + \name{} & Contriever & + \name{} & BM25 & + \name{} & BM25 & + \name{} \\ \midrule
NQ & 38.8 & \textbf{42.3}  & 47.4 & \textbf{50.8} & 47.6 & \textbf{50.4} & 40.9 & \textbf{42.4}  & 36.4$^{*}$ & \textbf{42.9}$^{*}$ \\
& {\small [37.3, 40.4]} & {\small [40.8, 43.7]}& {\small [45.9, 48.9]}& {\small [49.3, 52.4]}& {\small [46.2, 49.1]}& {\small [48.8, 51.8]}& {\small [39.5, 42.4]}& {\small [41.0, 43.9]}& {\small [32.8, 39.6]}& {\small [38.9, 46.8]}\\ 
WebQ & 32.5 & \textbf{36.6} & 37.5 & \textbf{40.4} & 37.7 & \textbf{40.9} & \textbf{36.4} & {32.1}  & 35.3$^{*}$ & \textbf{40.5}$^{*}$ \\ 
& {\small [30.8, 34.3]} & {\small [34.8, 38.5]}& {\small [35.8, 39.3]}& {\small [38.5, 42.3]}& {\small [35.8, 39.6]}& {\small [39.1, 42.7]}& {\small [34.5, 38.2]}& {\small [30.2, 34.0]}& {\small [32.2, 38.7]}& {\small [36.7, 43.9]}\\ 
2Wiki & 32.8 & \textbf{38.1} & 22.8 & \textbf{25.2} & 32.9 & \textbf{37.1} & 39.2 & \textbf{43.2}  & 31.2 & \textbf{36.2} \\ 
& {\small [28.9, 36.3]} & {\small [34.0, 42.0]}& {\small [19.4, 26.3]}& {\small [21.6, 28.8]}& {\small [29.3, 36.9]}& {\small [33.0, 41.4]}& {\small [35.2, 43.0]}& {\small [39.3, 47.2]}& {\small [27.7, 34.9]}& {\small [32.2, 39.9]}\\ 
HotpotQA & 40.3 & \textbf{43.4} & 34.6 & \textbf{35.5} & 42.9 & \textbf{43.5} & 44.4 & \textbf{50.4}  & \textbf{39.6} & {38.4} \\ 
& {\small [36.4, 44.4]} & {\small [39.4, 47.6]}& {\small [30.8, 38.1]}& {\small [31.6, 39.6]}& {\small [36.6, 47.2]}& {\small [39.3, 47.7]}& {\small [40.5, 48.1]}& {\small [46.3, 54.3]}& {\small [35.9, 43.2]}& {\small [34.6, 42.5]}\\ 
\bottomrule
\end{tabular}}
}
\label{table:supp_diff_llm_f1_ci}
}
\end{table*}



\end{document}